\documentclass[journal]{IEEEtran}

\ifCLASSINFOpdf
\else
\fi

\usepackage{float}
\usepackage[small]{caption}
\usepackage{subfig}
\usepackage{placeins}
\usepackage{graphicx}
\usepackage{amsmath}
\usepackage[ruled,vlined]{algorithm2e}
\usepackage[T1]{fontenc}
\usepackage{wrapfig}
\usepackage{tabularx}

\begin{document}

\title{Translational Symmetry-Aware Facade Parsing for 3D Building Reconstruction}
\author{Hantang~Liu,~Wentong~Li,~Jianke~Zhu~\IEEEmembership{Senior~Member,~IEEE}

\IEEEcompsocitemizethanks{\IEEEcompsocthanksitem Hantang Liu, Wentong Li and Jianke Zhu are with the College of Computer Science, Zhejiang University, Hangzhou, China, 310027. Jianke Zhu is also with the Alibaba-Zhejiang University Joint Research Institute of Frontier Technologies, Hangzhou, China.
	\IEEEcompsocthanksitem Jianke Zhu is the corresponding author (e-mail:jkzhu@zju.edu.cn).
}
}

\maketitle

\begin{abstract}
  Effectively parsing the facade is essential to 3D building reconstruction, which is an important computer vision problem with a large amount of applications in high precision map for navigation, computer aided design, and city generation for digital entertainments. To this end, the key is how to obtain the shape grammars from 2D images accurately and efficiently. Although enjoying the merits of promising results on the semantic parsing, deep learning methods cannot directly make use of the architectural rules, which play an important role for man-made structures. In this paper, we present a novel translational symmetry-based approach to improving the deep neural networks. Our method employs deep learning models as the base parser, and a module taking advantage of translational symmetry is used to refine the initial parsing results. In contrast to conventional semantic segmentation or bounding box prediction, we propose a novel scheme to fuse segmentation with anchor-free detection in a single stage network, which enables the efficient training and better convergence. After parsing the facades into shape grammars, we employ an off-the-shelf rendering engine like Blender to reconstruct the realistic high-quality 3D models using procedural modeling. We conduct experiments on three public datasets, where our proposed approach outperforms the state-of-the-art methods. In addition, we have illustrated the 3D building models built from 2D facade images.
\end{abstract}


%
\IEEEpeerreviewmaketitle

\section{Introduction}
\IEEEPARstart{C}reating 3D building models from 2D facade images has long been desired in the computer vision community, which has various applications in high precision map, computer aided design, and city generation for digital entertainments like movies and computer games. 

The key of automatic 3D building reconstruction is how to obtain the accurate facade parsing results, which is still a challenging problem. In this paper, we propose a novel approach to parsing building facades into shape grammars through deep learning model with translational symmetry. In addition, we make use of the procedural modeling to reconstruct 3D building models.
\begin{figure}[t]
  \centering
  \subfloat{\includegraphics[width=0.15\textwidth]{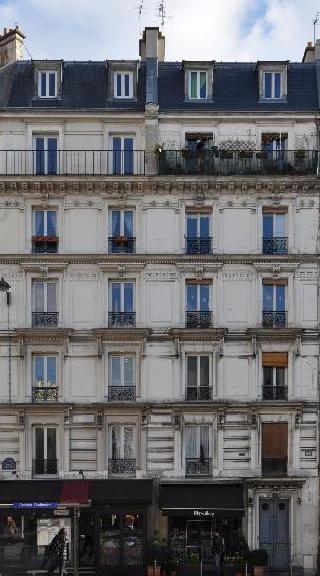}}
  \hspace{0.1em}
  \subfloat{\includegraphics[width=0.15\textwidth]{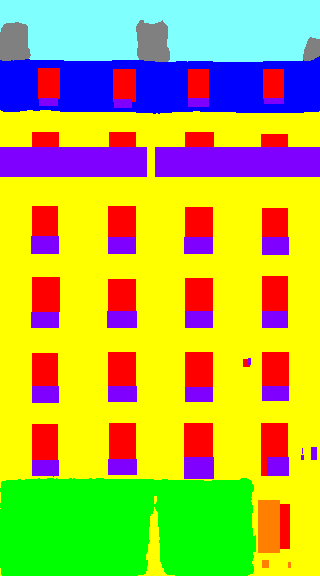}}
  \hspace{0.1em}
  \subfloat{\includegraphics[width=0.15\textwidth]{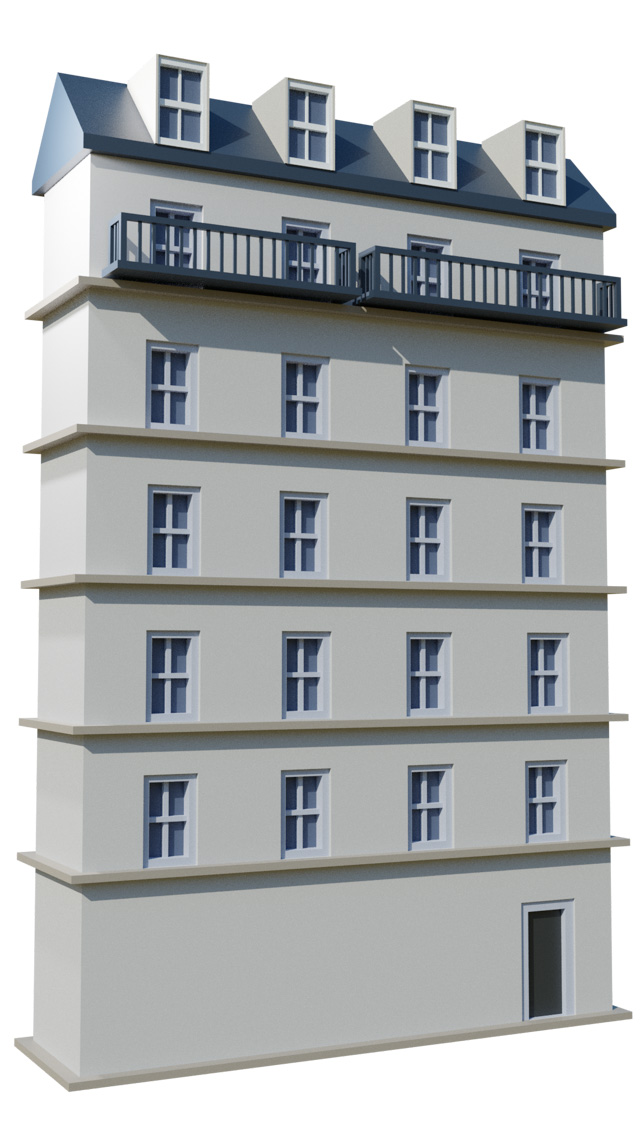}}
  \caption{The pipeline of our proposed approach. We first parse the facade into semantic labels, and then reconstruct the 3D building model using procedural modeling.}
  \label{fig:intro-demo}
  \end{figure}

The conventional approaches~\cite{teboul2010segmentation,martinovic2012three} usually employ the predefined shape grammars, and treat the facade parsing as template parameters estimation problem. On the other hand, deep learning models~\cite{liu2020deepfacade,DBLP:journals/cgf/NishidaBA18} learn the rules from input images. Most of these methods have formulated the facade parsing as either semantic segmentation or bounding box prediction problem. As deep learning models have already shown their outstanding performance in various computer vision tasks, we also take the deep learning approach in this paper. 

Generally, architectural rules play an essential role in the facade parsing process due to the strong prior information in man-made structures. One of them is the symmetry rule. There are two typical kinds of symmetries in modern buildings, including the reflective and translational symmetry. A single object is usually with reflectively symmetric, such as windows and doors. Translational symmetry means that all objects of the same kind in a facade have the similar size and appearance. 

Since neural networks tend to learn the underlying mapping from the training data, it is an open problem on how to add the prior information into the training and inference process. Liu et al.~\cite{liu2020deepfacade} enforce the reflective symmetry to train the neural network, which has improved both the parsing accuracy and visual quality. However, the reflective symmetric loss has two major drawbacks. One is that the symmetry is treated as a weak constraint. It only guides the neural network towards the reflective symmetry while cannot guarantee the good parsing results. The other is that the reflective symmetry constraint only works for a single object. The geometry is optimized by its own, which does not take its neighborhood into consideration.

To address the above issues, we propose a translational symmetry-based refinement module for neural networks in this paper. Differently from Liu et al.~\cite{liu2020deepfacade}, we optimize the output from deep neural network. This imposes the translational symmetry as a strong constraint, which leads to more accurate and visually satisfying results. To facilitate the effective facade parsing, we propose a novel deep learning model fusing semantic segmentation with anchor-free detection~\cite{DBLP:journals/corr/abs-1904-07850,DBLP:conf/cvpr/ZhouZK19,DBLP:conf/cvpr/RedmonDGF16}. In contrast to the two stage neural networks~\cite{liu2020deepfacade}, the proposed approach does not rely on a region proposal network, which makes it easier to train and faster to inference. Moreover, it is convenient for the anchor-free approach to incorporate different branches like segmentation and object attributes. Therefore, our proposed neural networks is able to use different kinds of information including segmentation map, bounding boxes and object attributes, which outperforms a typical segmentation network with bounding box prediction~\cite{liu2020deepfacade}.

Once the parsing results are obtained, we employ an off-the-shelf rendering engine like Blender~\cite{blender} to reconstruct the realistic high-quality 3D models using procedural modeling, as shown in Fig.~\ref{fig:intro-demo}. We demonstrate the efficacy of our method by comparing the state-of-the-art methods quantitatively on three popular facade parsing datasets. Their visual results show that our proposed method is promising for the semantic building reconstruction.

\section{Related Work}
Our work is related to facade parsing~\cite{teboul2010segmentation}, 3D building reconstruction~\cite{DBLP:journals/tog/MullerWHUG06} and deep learning~\cite{DBLP:conf/nips/KrizhevskySH12}. In general, facade parsing can be categorized into two groups. Conventional methods rely on the user-defined shape grammar rules. On the other hand, the deep neural network-based approaches tend to directly learn these rules from data. As for 3D building reconstruction, the most popular approach is to conduct the procedural modeling from shape grammars~\cite{DBLP:journals/tog/MullerZWG07}\cite{DBLP:journals/tog/MullerWHUG06}, which highly depends on the quality of parsing results. Thus, the facade parsing is key to the high quality 3D building models. We will look into these methods in the following.

\subsection{Conventional Facade Parsing Methods}

Most of conventional facade parsing methods relied on the predefined procedural grammars, which aim at finding the best parametric setting of a facade. It is easy for them to incorporate the architectural rules into the inference process while they are usually computationally demanding and tend to fail in cluttered environments~\cite{DBLP:conf/cvpr/MartinovicKRG15,DBLP:conf/cvpr/KozinskiGZOM15}.

Zhao et al.~\cite{DBLP:conf/cvpr/ZhaoFXZZQ10} assume most of vertical lines belong to buildings, where the input image are parsed into building, sky and ground. Then, they further refine the extracted shape of each facade unit. Wendel et al.~\cite{DBLP:conf/dagm/WendelDB10} and Recky et al.~\cite{DBLP:conf/3dim/ReckyWL11} try to find the repetitive patterns in facades to do the parsing. Koutsourakis et al.~\cite{DBLP:conf/iccv/KoutsourakisSTTP09} build the models through a set of basic shapes with the parametric rules. A tree representation of variable depth and complexity is used to account for the elaborate and varying architectural styles. They employ Markov Random Field~(MRF) to optimize the parameters of procedural grammars for a building facade. Ripperda and Brenner~\cite{ripperda2006reconstruction} also employ a tree representation, where the reversible jump Markov Chain Monte Carlo is used in the tree construction. Teboul et al.~\cite{teboul2011shape} formulate the shape grammars using reinforcement learning, where the promising results can be achieved if buildings conform to these grammars.

Mathias et al.~\cite{DBLP:journals/ijcv/MathiasMG16} propose a three-layer approach to facade parsing. They hand-crafted the prior knowledge into each layer. With the first layer, a recursive neural network (RNN) is trained to label facades at super-pixel level. In the middle layer, they introduce the knowledge about distinct facade elements. Then, they combine the output of RNN with object detectors, and treat the merging procedure as a 2D Markov Random Field over the pixels. Cohen et al.~\cite{Cohen_2014_CVPR} formulate the facade parsing as a sequential optimization problem instead of the usual classification task or grammar learning, where dynamic programming is used to solve the optimization problem.  

\subsection{Deep Learning-based Methods}
There are a few research efforts having been devoted to tackling the facade parsing problem using deep learning techniques. DeepFacade~\cite{DBLP:conf/ijcai/LiuZZH17,liu2020deepfacade} suggest a reflective symmetric loss as a constraint to train the neural networks. In~\cite{DBLP:conf/ijcai/LiuZZH17}, Liu et al. propose a variance-based loss, where the centers of horizontal and vertical line segments should lie on the same straight line for symmetric objects. Later, they employ a two-stage object detector~\cite{liu2020deepfacade}, in which the predicted bounding boxes are treated as the symmetry indicator. Schmitz and Mayer~\cite{schmitz2016convolutional} train CNN on images patches to parse the facade. John Femiani et al.~\cite{DBLP:journals/corr/abs-1805-08634} propose three different network architectures to perform multi-label facade image segmentation, where each one has a unique feature.


Our approach is also related to image segmentation, scene parsing and object detection. Long et al.~\cite{long2015fully} are the first to train an end-to-end deep convolutional neural network for general image segmentation task. Chen et al.~\cite{DBLP:journals/corr/ChenPKMY14}\cite{DBLP:journals/corr/ChenPSA17}\cite{DBLP:journals/pami/ChenPKMY18} employ the dilated convolution instead of plain convolution. This approach avoids the use of a deconvolution layer, and make the network easier to train. CRF post-processing~\cite{DBLP:journals/corr/ChenPKMY14} can be applied to refine the results. Encoder-decoder models~\cite{DBLP:journals/pami/BadrinarayananK17}, feature pyramids~\cite{DBLP:conf/cvpr/LinDGHHB17}\cite{DBLP:conf/cvpr/ZhaoSQWJ17} and skip-connection~\cite{DBLP:conf/miccai/RonnebergerFB15} are also important techniques in semantic segmentation. Abdulnabi et al.~\cite{DBLP:journals/tmm/AbdulnabiSZCW18} propose a multi-modal recurrent neural networks (RNNs) for indoor scene labeling. They train two RNNs at the same time, and connect the two networks through an information transfer layer. The recent object detection techniques can be grouped into two stage approach and single stage method. Two stage approaches~\cite{ren2015faster,DBLP:conf/iccv/Girshick15,DBLP:conf/iccv/HeGDG17} first employ a region proposal network to find the potential bounding boxes, and then predict their labels via a classifier. Single stage methods~\cite{DBLP:conf/cvpr/RedmonDGF16,DBLP:conf/eccv/LiuAESRFB16} do not rely on the region proposal networks.~\cite{DBLP:journals/tmm/LiLLWXFY18} combine the object detection and semantic segmentation. Zhou et al.~\cite{DBLP:journals/corr/abs-1904-07850,DBLP:conf/cvpr/ZhouZK19} represent the objects as points with the extendable attributes. For example, they represent bounding boxes as the width and height from the center point, or the corner location of a bounding box.




\begin{figure*}[t]
  \centering
  \includegraphics[width=0.99\textwidth]{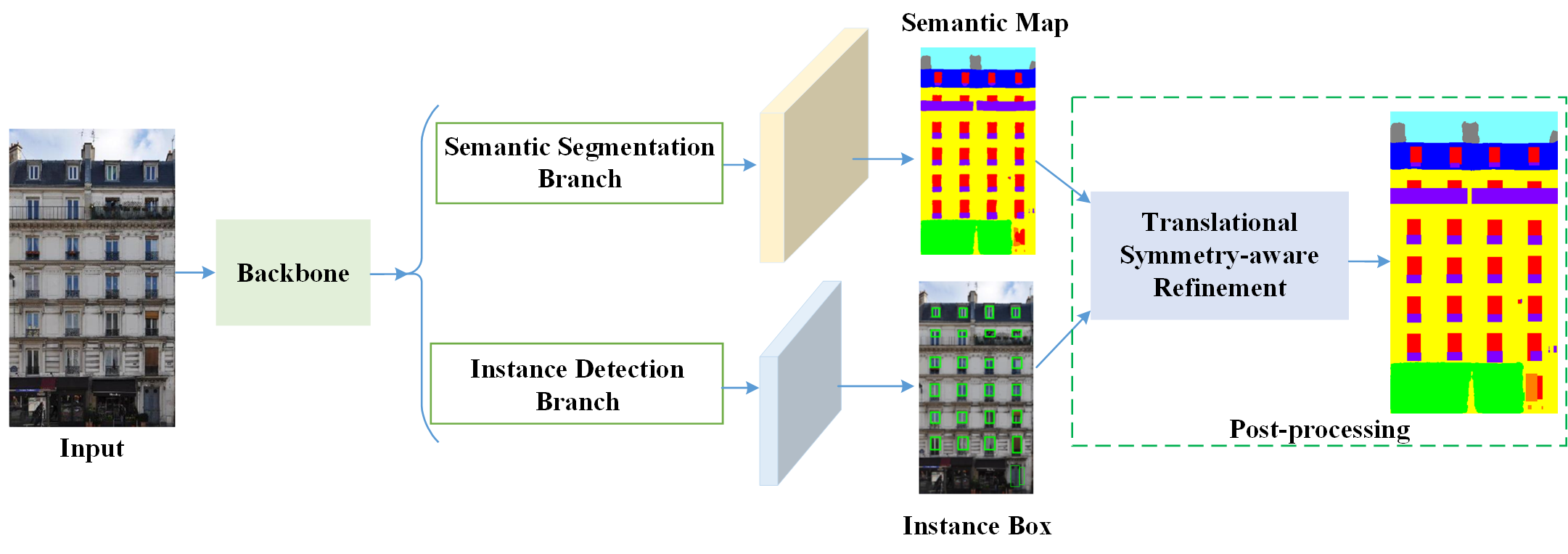}
  \caption{Our proposed network architecture consists of two branches. One is for semantic segmentation, another one branch is for instance detection based anchor-free framework. The facade refinement are performed by the translational symmetry during the post-progressing.}
  \label{fig:network-structure}
\end{figure*}

\subsection{3D Building Reconstruction}
Usually, the 3D buildings are represented by either the point cloud or procedural models. The former relies on the geometric information while the latter focuses on the semantics.

A mature pipeline for reconstructing 3D urban scenes from images is generally based on structure-from-motion techniques~\cite{DBLP:conf/eccv/BrostowSFC08}\cite{DBLP:journals/ijcv/LadickySRSBCT12}\cite{DBLP:conf/eccv/MunozBH12}\cite{DBLP:conf/cvpr/AnguelovTCKGHN05}. Their output are typically point clouds or meshes without the labels like windows and doors, however, the semantically reconstructed 3D models are needed in real-world applications. Martinovic et al.~\cite{DBLP:conf/cvpr/MartinovicKRG15} employ CRF to predict the labels of point clouds and meshes in order to semantically classify the facade elements.

Muller et al.~\cite{DBLP:journals/tog/MullerZWG07,DBLP:journals/tog/MullerWHUG06} propose a high quality procedural modeling method, which is able to generate the high quality 3D building models with the predefined shape grammars. However, the parsing capability of conventional approaches is limited comparing to the deep learning-based methods. Nishida et al.~\cite{DBLP:journals/cgf/NishidaBA18} propose an interactive tool to extract the building grammar from a single image, where the user needs to mark out the silhouette. Convolutional neural networks are used to extract different components of a facade from large-scale building mass to fine-scale windows and doors geometry. Their method generates visually satisfied results, which mainly depends on a window detector with the predefined shape components allowing some deviations. In this paper, our presented method serves as a general facade parser that aims to recovering the accurate layout of building facades.

\section{Anchor-free Facade Parsing}
In this section, we first present our proposed anchor-free facade parsing framework, and then give the formulation on both segmentation branch and detection head. Finally, we describe the details on training and inference.
\subsection{Neural Network Architecture}
 The conventional two-stage methods~\cite{ren2015faster} first compute potential regions using a region proposal network, and then extract the features from neural network. Moreover, each region proposal is further classified. Therefore, two-stage methods essentially formulate the detection as a region proposal classification problem.
 
 In contrast to the two stage scheme~\cite{DBLP:conf/ijcai/LiuZZH17,liu2020deepfacade}, we propose a novel anchor-free neural network architecture by fusing the semantic segmentation and detector within the same framework. Moreover, anchor-free approach is more like semantic segmentation rather than two-stage proposal classification, which can not only reduce the computational cost but also enjoys the merits of better convergence.


In this paper, we choose FRRN~\cite{DBLP:conf/cvpr/PohlenHML17} as the backbone network, since it does not require to be pre-trained on a large dataset like ImageNet~\cite{deng2009imagenet}. Thus, we can learn the neural network model from scratch solely using the facade dataset while still obtaining the promising results. The FRRN is a typical encoder-decoder architecture network, where the encoder part downscales the feature map 16 times with respect to the original image $I\in R^{W\times H\times 3}$. Moreover, the consecutive transposed convolution is employed to upsample the feature map.

Our network has two branches, as shown in Fig.~\ref{fig:network-structure}. One branch is for semantic segmentation, and another is for anchor-free detection. We aim to predicting a semantic segmentation response map with the keypoint detection map. The former is a full sized map as the original image while the latter is down-scaled with a stride of 4. Thus, the training loss is made of five different terms as follows:
\begin{equation}
L = L_{ce} + \lambda_1 L_{det} + \lambda_2 L_{wh} + \lambda_3 L_{off} + \lambda_4 L_{corner}
\end{equation}
where $L_{ce}$ is the segmentation loss, $L_{det}$ is the focal loss for detection, $L_{wh}$ is the size loss, $L_{off}$ is the local offset loss, and $L_{corner}$ is the corner position loss. $\lambda_1$, $\lambda_2$, $\lambda_3$ and $\lambda_4$ are the weights for each part of the loss, respectively. Since all the losses are normalized in our implementation, we set $\lambda_1 = \lambda_2 = \lambda_3 =\lambda_4 = 1$, empirically.

\subsection{Semantic Segmentation Branch}
For the semantic segmentation branch, we use the cross-entropy loss function~\cite{long2015fully}:
\begin{equation}
  \label{eq:cross-entropy}
  L_{ce}(\mathbf{x}, \mathbf{y}) = -\sum_{c}^{M} \sum_{i}^{N}y_{i,c}\log (p_{i,c})
\end{equation}
where $\mathbf{x}$ is the input image array, and $\mathbf{y}$ is the probability distribution of the category label of the image. $M$ is the number of classes of the dataset, and $N$ is the number of pixels in the image. $i$ is pixel index. $y_{i,c}$ is the true probability distribution, and $p_{i,c}$ is the prediction.

The semantic segmentation branch is an encoder-decoder architecture that restores the predicted heatmap to the original image size, instead of downsampling by a factor of 4 as the detection branch. This is because semantic segmentation requires a finer heatmap in order to obtain the better results. If the predicted heatmap is downsampled, then the enlarged final results will be not accurate enough.


\subsection{Detection Branch}
Anchor-free methods~\cite{DBLP:journals/corr/abs-1904-07850} detect objects by keypoint map, where each pixel represents the potential of being the center for an object. This is more like a segmentation response map rather than a classification task setting. Therefore, it is natural to fuse segmentation with anchor-free object detection.

For the detection branch, we employ the base structure of CenterNet~\cite{DBLP:journals/corr/abs-1904-07850}. We represent each object in a facade as the center point, whose size is an attribute of each point. Besides predicting the size, we employ an auxiliary corner point detection in the training process. The output stride downsamples the output prediction by a factor $R = 4$.
Such scheme is able to reduce computational cost while retaining the high accuracy.

The training objective is a focal loss~\cite{DBLP:conf/cvpr/LinDGHHB17}:
\begin{align}
L_{det} = -\frac{1}{N}\sum_{i=1}^H\sum_{j=1}^W
\begin{cases}
(1-\hat Y_{ij})^\alpha \log(\hat Y_{ij}) \quad \quad  \quad   \ \ if \ Y_{ij}=1 \\
(1-Y_{ij})^\beta(\hat Y_{ij})^\alpha \log(1-\hat Y_{ij}) \quad \  o.w.
\end{cases}
\end{align}
where $N$ is the number of instances in image. $\hat Y_{ij}\in [0,1]$ denote the predicted probability for the position $(i,j)$. $\alpha$ and $\beta$ are the hyper-parameters.

The object size is predicted by estimating the width and height of the bounding box. $(x_1, y_1, x_2, y_2)$ is denoted as the bounding box of an object with the size of $(w, h)$, where $w = x_2 - x_1$ and $h = y_2 - y_1$. Therefore, we employ an L1 regressor to predict the size for each object:
\begin{equation}
L_{wh} = \frac{1}{N}\sum_{k=1}^N |\hat{w} + \hat{h} - (w + h)| 
\end{equation}

During the inference, discretization error usually occurs due to the offset drifting. To tackle this issue, we predict a local offset $\hat O\in \mathcal{R}^{\frac{W}{R}\times \frac{H}{R}\times 2}$ for each center point with the following loss function:
\begin{equation}
L_{off} = \frac{1}{N}\sum_p|\hat O_{\tilde p}-(\frac{p}{R}-\tilde p)|
\end{equation}
where $p \in \mathcal{R}^2 $  denotes the center point of  ground truth, and  $\tilde p $ is computed with $\lfloor \frac{p}{R} \rfloor$.

We can add corner points as the additional attribute of an object. Although we only need to predict the location and size of the window in a parametric setting, estimating the corner location still greatly improves the training process. Such scheme provides the extra supervision information for the network during training, which leads to better convergence and promising prediction results. Moreover, corners are not needed in the inference stage, which do not incur the extra computational cost in prediction.

As the previous offset head~\cite{DBLP:journals/corr/abs-1904-07850}, we only need to add a few channels to represent each corner of the object, where four corners are estimated for each subject with a feature map $\hat C\in \mathcal{R}^{\frac{W}{R}\times \frac{H}{R}\times 8}$. Each point takes two channels, and there are four corners for each object center point. Thus, we have 8 additional channels in total. Then, the corner loss can be derived as follows:
\begin{equation}
L_{corner} = \frac{1}{N}\sum_q|\hat C_{\tilde q}-(\frac{q}{R}-\tilde q)|
\end{equation}
where $\tilde q= \lfloor \frac{q}{R} \rfloor$.
The loss only occurs at each corner $q$, and other locations are ignored in our implementation.

\section{Translational Symmetry-aware Refinement}
In this section, we first define the translational symmetry in building facades, and then describe the evaluation metrics for translational symmetry. Finally, we present a novel facade refinement method based on translational symmetry constraints.

\subsection{Motivation}
\begin{figure}[h]
  \centering
  \includegraphics[width=0.48\textwidth]{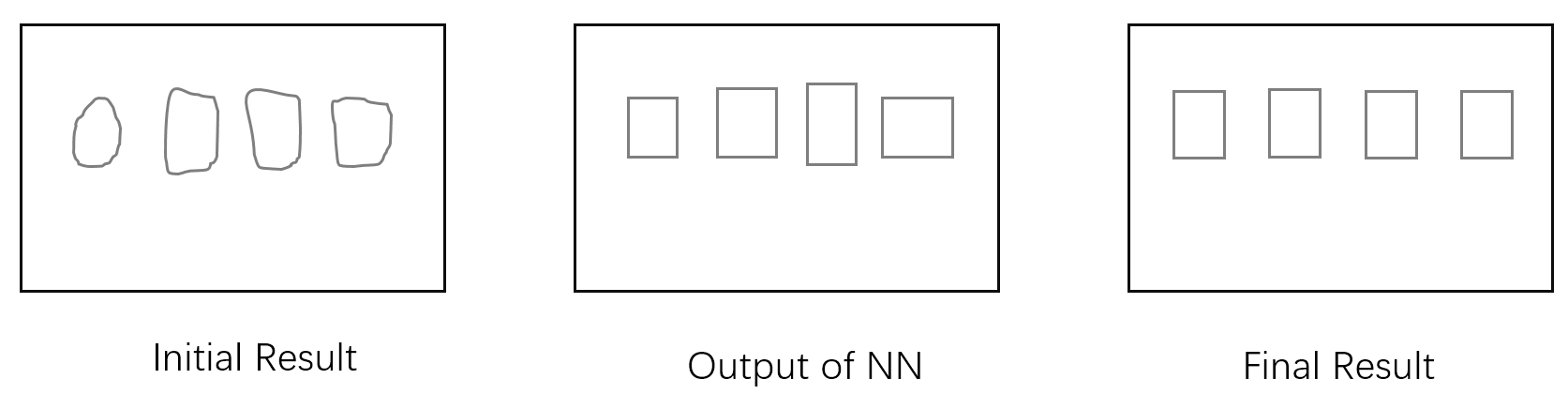}
  \caption{Motivation for the facade objects refinement using translational symmetry.}
  \label{fig:motivation}
\end{figure}

In general, translational symmetry exists almost in every facade of modern urban buildings. This means that the property of an object stays invariant under translation. For a building facade, the same kind of objects are with the similar appearance and size. In particular, objects of the same category share the same shape and size, whose centers usually fall on a straight line. Typically, translational symmetry in facades have a direction of either horizontal or vertical. A facade is translational symmetric in horizontal means that the elements are same in the horizontal direction. It is also true for the vertical direction. Some facades are translationally symmetric in both directions, while others are only symmetric in one direction. This makes it necessary for us to decide in which direction the building is symmetric. Thus, we need to know how symmetric a facade is.

\subsection{Translational Symmetry Metric}
We define a quantitative metric to evaluate the symmetry of a facade. For building facades, translational symmetry relies on two aspects. Firstly, the objects are translational symmetric, whose center coordinates have equal distances in the symmetric orientation and share the same value in the orthogonal direction. Secondly, the sizes are invariant in the symmetric direction. Based on these assumptions, the translational symmetry metric $T$ is composed of two terms:
\begin{equation}
  T = T_{c} + T_{s}
\end{equation}
where $T_{c}$ represents the score of how symmetric the centers are, and $T_{s}$ is the score of how symmetric the sizes are. Since $T$ has an orientation, we denote $T^{h}$ as the measure for horizontal direction and $T^{v}$ for vertical one.

We take horizontal translational symmetry as an example, which should satisfy two conditions: 1) their vertical coordinates need to be as close as possible; 2) the distances between horizontal coordinates should be the same. Therefore, we derive the following formulation for $T_{c}$:
\begin{equation}
  T_{c} = \frac{1}{N}\sum_{i=1}^N (y_i - \hat{y})^2 + \frac{1}{N-1}\sum_{i=1}^{N-1}(\Delta x_i - \hat{\Delta x})
\end{equation}

For $T_{s}$, all objects may have the same size. Thus, the bigger the variance of sizes, the less the symmetry is. We formulate $T_{s}$ as below:
\begin{align}
  T_{s} & = \frac{1}{N}\sum_{i=1}^N (\mathbf{s}_i - \hat{\mathbf{s}})\cdot (\mathbf{s}_i - \hat{\mathbf{s}}) \\
           & = \frac{1}{N}\sum_{i=1}^N [(w_i - \hat{w})^2 + (h_i - \hat{h})^2]
\end{align}
where $\mathbf{s}_i=(w_i,h_i)$ represents the width and height of object $i$, and $N$ is the number of objects.

\subsection{Facade Refinement by Translational Symmetry}
Based on the above formulation, we find that the lower $T$ leads to the higher degree of translational symmetry. We aim at solving the following minimization problem:
\begin{equation}
\mathop {\operatorname{argmin} }\limits_{\mathbf{x},\mathbf{y},\mathbf{S}} T = \mathop {\operatorname{argmin}}\limits_{\mathbf{x},\mathbf{y},\mathbf{S}} ({T_c} + {T_s})
\end{equation}
where $\mathbf{x}=\{x_1,x_2,\dots,x_N\}$ are the horizontal coordinates of all the objects, $\mathbf{y}=\{y_1,y_2,\dots,y_N\}$ denote all the vertical coordinates, and $\mathbf{S}=\{\mathbf{s}_1,\mathbf{s}_2,\dots,\mathbf{s}_N\}$ represents all the sizes. Our goal is to find a configuration to minimize $T$.

The straightforward solution is to set $x_i=\hat x,y_i=\hat y, \mathbf{s}_i=\hat{\mathbf{s}}$, however, which has some issues. As the facade may not always be symmetry in both directions, we have to estimate the symmetric direction first. Assuming that a facade has symmetry in one direction and use it as the refinement target. If $T^{h}<T^{v}$, we apply the horizontal refinement. Otherwise, we refine the vertical direction. Another problem is that the windows may have slightly different sizes even in the symmetric direction. Practically, we introduce a weighted sum to allow one object to keep the size of its initially predicted result.
\begin{align}
  x_{new} = \tilde{T}\cdot x_i + (1-\tilde{T})\hat{x}\\
  \mathbf{s}_{new} = \tilde{T}\cdot \mathbf{s}_i + (1-\tilde{T})\hat{\mathbf{s}}
\end{align}
where $\tilde{T}=\sigma(T)\in [0,1]$ is the result of $T$ being scaled to $[0,1]$, and $\sigma$ is the sigmoid function. $x_{new}$ is the new center locations, and $\mathbf{s}_{new}$ are sizes of objects after refinement. $x_{i}$ and $\mathbf{s}_{i}$ are the corresponding values before refinement, $\hat{x}$ and $\hat{\mathbf{s}}$ are the mean.

\section{Experiments on Facade Parsing}
In this section, we conduct the intensive experiments on three popular datasets to demonstrate the efficacy of our proposed anchor-free facade parsing approach. Moreover, we not only provide the quantitative evaluation but also show the facade parsing results.

\begin{table*}[t!]
	\centering
	\begin{tabular}{|c|c|c|c|c|c|c|c||c|c|c|c|c|c|}
		\hline
		Class  & \protect\cite{DBLP:journals/ijcv/MathiasMG16}\textsuperscript{1} & \protect\cite{DBLP:journals/ijcv/MathiasMG16}\textsuperscript{2} & \protect\cite{Cohen_2014_CVPR}\textsuperscript{1} &
		\protect\cite{Cohen_2014_CVPR}\textsuperscript{2} &
		\protect\cite{Cohen_2014_CVPR}\textsuperscript{3} & \cite{yang2011regionwise} & \cite{DBLP:journals/corr/abs-1805-08634} &  FCN-8s  & DeepFacade~\cite{liu2020deepfacade} & DeepFacade~\cite{liu2020deepfacade}\textsuperscript{IOU} & Ours & Ours\textsuperscript{IOU} \\
		\hline
		Window  & 76 & 78 & 68 & 87 & 85 & 62 & 95.6 & 86.8 & 97.6 & 80.3 & 97.8 & 82.5\\
		\hline
		Wall  & 90 & 89 & 92 & 88 & 90 & 82 & 91.7 & 96.0 & 97.9 & 89.8 & 97.5 & 91.6\\
		\hline
		Balcony  & 81 & 87 & 82 & 92 & 91 & 58 & 96.0 & 92.4 & 96.2 & 85.2 & 96.3 & 87.2\\
		\hline
		Door  & 58 & 71 & 42 & 82 & 79 & 47 & 98.8 & 86.0 & 92.3 & 63.1 & 97.6 & 70.6\\
		\hline
		Roof  & 87 & 79 & 85 & 92 & 91 & 66 & 97.7 & 92.7 & 97.7 & 75.6 & 97.3 & 76.2\\
		\hline
		Sky  & 94 & 96 & 93 & 93 & 94 & 95 & 98.9 & 96.6 & 98.2 & 84.2 & 98.0 & 88.4\\
		\hline
		Shop  & 97 & 95 & 94 & 96 & 94 & 88 & 98.4 & 95.6 & 96.0 & 80.3 & 95.4 & 84.3\\
		\hline
		Chimney  & - & - & 54 & 90 & 85 & - & 96.9 & 85.3 & 90.5 & 64.6 & 90.8 & 68.5\\
		\hline
		total acc.  & 88.0 & 88.0 & 86.7 & 89.9 & 90.3 & 74.7 & 96.7 & 93.7 & 97.3 & 77.9 & 97.1 &81.2\\
		\hline
	\end{tabular}
	\caption{Comparisons on ECP dataset(\%). \protect\cite{DBLP:journals/ijcv/MathiasMG16}\textsuperscript{1} and \protect\cite{DBLP:journals/ijcv/MathiasMG16}\textsuperscript{2} denote two variants of \protect\cite{DBLP:journals/ijcv/MathiasMG16}. \protect\cite{Cohen_2014_CVPR}\textsuperscript{1}, \protect\cite{Cohen_2014_CVPR}\textsuperscript{2}, and \protect\cite{Cohen_2014_CVPR}\textsuperscript{3} are three variants of \protect\cite{Cohen_2014_CVPR}. FCN-8s directly employ Fully Convolutional Networks~\cite{long2015fully} on ECP dataset. Then, we compare our results with DeepFacade on both pixel accuracy and IOU.}
	\label{tab:ecp-compare}
\end{table*} 

\subsection{Experimental Setup}
 To facilitate the fair comparison against the state-of-the-art methods, we evaluate our presented method on three typical facade datasets, including the Ecole Central Paris (ECP) dataset~\cite{ecpdataset}, the RueMonge dataset\cite{DBLP:conf/cvpr/MartinovicKRG15}, and the ArtDeco dataset~\cite{gadde2016learning}. 

The ECP dataset is a popular 2D facade parsing testbed with a total number of 104 annotated building images. All annotations are pixelwise-labeled semantic segmentation maps. This dataset has 8 classes, including window, wall, balcony, door, shop, sky, chimney and roof. All the images in ECP dataset contain the rectified and cropped facades of Haussmannian style buildings in Paris. The original annotation labels the images using a Haussmannian-style grammar. This often results in the imprecise or even wrong annotations. Therefore, we employ the labels provided by \cite{teboul2010segmentation}, where the annotations can better fit the ground truth. Some methods~\cite{schmitz2016convolutional} only focus on windows and walls, which do not use full annotations. In this paper, we make full use of all the labels.

The Ruemonge dataset has 428 facade images in total, where only a portion of them are annotated. The training set has 113 images, and the testing split contains 202 images. For a single building, it may have several pictures from slightly different perspective. This is because the images are taken from a camera moving down a street. Thus, overlapping occurs frequently. Differently from ECP dataset, Ruemonge dataset has 3D point clouds and mesh annotations. Previous method~\cite{DBLP:conf/cvpr/MartinovicKRG15} has explored both 2D and 3D information, where the 3D method runs much faster than the 2D approach with the higher accuracy. Our method is a pure 2D approach.

The ArtDeco dataset consists of 80 rectified facade images with similar style. In contrast to other datasets, some of images have occlusions due to vegetation. This makes it ideal for testing the inference of architectural elements in the presence of large occlusions, as it provides the hand-annotated ground truth for the labels behind vegetation.

We train our network from scratch on each dataset. Our proposed framework has single neural network with multiple heads. Such scheme is much easier to train comparing against DeepFacade~\cite{liu2020deepfacade}, where they have to train two networks simultaneously. Usually, our training process is able to converge within 140 epochs. We conducted our experiments on a PC with 4-GPU at a initial learning rate of 2.5e-4.

\subsection{Annotations for Facade Object Detection}

As the facade datasets only provide the semantic segmentation labeling, i.e., heatmap for the images, we have to generate object detection annotations from these labels. We employ the automated labeling method in order to avoid introducing new supervision in the experiments, which may not be fair to other methods for comparisons.

We manage to generate the bounding boxes for windows, balconies and doors by extracting each facade object instance from the semantic segmentation map. We find that all pixels in the same object are connected, and all the different connected regions are objects. Thus, we compute the convex hull of the object region, and then find the minimal bounding box containing this convex hull. Therefore, we can easily generate the bounding boxes for each facade objects from the semantic segmentation map.

For the four corner locations of each object, we find them through a sorting algorithm. If we set the top left corner of the entire image as the origin, then the top left corner of object is the point having the least distance to the origin in an object. Similarly, the other three points can be found by setting the origin to the top-right, bottom-right and bottom-left corners of the image. Thus, we can directly extract the corner locations from the segmentation map without resorting to the manual labeling.

\subsection{Quantitative Evaluation}

We conduct the quantitative experiments on three datasets to demonstrate the efficacy of our method.

\subsubsection{Evaluation metrics} Two performance metrics are used to evaluate the facade parsing results, including Pixel accuracy and Jaccard Index. The latter is also known as Intersection Over Union (IOU), which is a more reasonable and widely used metric for evaluating semantic segmentation results. Pixel accuracy is defined as follows:
\begin{equation}
pixel\_acc = \frac{TP}{TP+FN}
\end{equation}
and Jaccard Index (per class) is defined as below:
\begin{equation}
IOU_c = \frac{TP}{TP+FP+FN}
\end{equation}
where TP means true positive, FN denotes false negative and FP is false positive. The mIOU is the average of the IOUs for all classes:
\begin{equation}
mIoU = \frac{1}{N}\sum_{i=1}^N IOU_c
\end{equation}

Although pixel accuracy is a popular evaluation metric in the previous facade parsing methods~\cite{DBLP:journals/ijcv/MathiasMG16, Cohen_2014_CVPR,yang2011regionwise,DBLP:journals/corr/abs-1805-08634,liu2020deepfacade}, it is easy to incur overfitting of one class, especially for those classes with large weight. On the other hand, Jaccard index can better reflect the prediction results. This is because high pixel accuracy usually has high recall, yet not necessarily high accuracy for all classes. The class with more pixels play a more important role in the final pixel accuracy. Thus, it is unfair for other classes. As pixel accuracy does not account for the false positives, this metric tends to over-label positives, which is not the optimal results. We prefer to IOU metric in our experiments. Moreover, we report pixel accuracy in order to compare the results with other methods, which rarely provide IOU scores.

\subsubsection{Results on ECP dataset} Table~\ref{tab:ecp-compare} reports the results of different approaches on the ECP dataset. Most of previous methods report the pixel accuracy on this dataset. DeepFacade~\cite{liu2020deepfacade} is the state-of-the-art approach, which is the only method having reported IOU scores. It can be observed that our presented method has achieved higher overall IOU on the ECP dataset than DeepFacade. Moreover, we outperform the state-of-the-art methods on each class, especially on those important categories like window, balcony and door, where the significant improvement have been achieved.


\begin{table}[h]
	\centering
	\begin{tabular}{|c|c|c|c|c|}
		\hline
		Class & ~\cite{DBLP:conf/cvpr/MartinovicKRG15} 2D & ~\cite{DBLP:conf/cvpr/MartinovicKRG15} 2D+3D & DeepFacade~\cite{liu2020deepfacade} & Ours\\
		\hline
		mIOU & 57.53 & 63.32&63.78 &65.35\\
		\hline
	\end{tabular}
	\caption{Quantitative results on the RueMonge dataset(\%). We compare our method against Martinovic~\cite{DBLP:conf/cvpr/MartinovicKRG15} and DeepFacade~\cite{liu2020deepfacade}. For Martinovic~\cite{DBLP:conf/cvpr/MartinovicKRG15}, we compare with both the 2D and 2D+3D setting.}
	\label{tab:chap5-ruemonge}
\end{table}

\subsubsection{Results on Ruemonge dataset} Table~\ref{tab:chap5-ruemonge} shows the experimental results on Ruemonge dataset. We compare our proposed approach with two state-of-the-art methods, including DeepFacade~\cite{liu2020deepfacade} and Martinovic et al~\cite{DBLP:conf/cvpr/MartinovicKRG15}. Martinovic et al.~\cite{DBLP:conf/cvpr/MartinovicKRG15} have several different settings. As this dataset has both 2D and 3D data, they have reported mIOU scores obtained by reasoning in 2D or 3D spaces only and combining both 2D and 3D results together. Their best accuracy is obtained by combining 2D and 3D output. Our presented method only make use of the 2D image data and do not exploit the 3D scatter points. The experimental results demonstrate that our proposed 2D method has outperformed the 2D method in~\cite{DBLP:conf/cvpr/MartinovicKRG15} at a large margin. Moreover, we obtain better results than the 2D+3D method using point cloud. Additionally, our proposed approach perform better than the recent DeepFacade method~\cite{liu2020deepfacade} on this dataset. 

\begin{table*}[ht]
	\centering
	\begin{tabular}{|c|c|c|c||c|c|c|c|}
		\hline
		Class & Cohen~\cite{DBLP:conf/cvpr/CohenSP14} & Kozinski~\cite{DBLP:conf/cvpr/KozinskiGZOM15} & Cohen~\cite{DBLP:conf/3dim/CohenOLP17} & DeepFacade~\cite{liu2020deepfacade} & DeepFacade~\cite{liu2020deepfacade}\textsuperscript{IOU} & Ours & Ours\textsuperscript{IOU}\\
		\hline
		Roof & 84 & 82 & 85 & 83.4 & 72.8 & 83.9 & 79.3 \\
		Shop & 97 & 97 & 97 & 97.6 & 92.3 & 97.7 & 93.7\\
		Balcony & 85 & 87 & 86 & 87.3 & 64.6 & 88.2 & 71.8\\
		Sky & 94 & 97 & 95 & 96.8 & 93.8 & 96.9 & 93.6\\
		Window & 82 & 82 & 82 & 95.4 & 70.7 & 96.2 & 80.1\\
		Door & 56 & 57 & 65 & 98.7 & 82.3 & 98.7 & 85.0\\
		Wall & 88 & 88 & 88 & 84.7 & 82.2 & 86.0 & 82.2\\
		\hline
		Overall & 85.3 & 88.8 & 88.3 & 92.9 & 79.8 & 93.5 & 83.7\\
		\hline
	\end{tabular}
	\caption{Results on the ArtDeco dataset(\%). Numbers represent pixel accuracy except DeepFacade\textsuperscript{IOU} and Ours\textsuperscript{IOU}. We compare with Cohen~\cite{DBLP:conf/cvpr/CohenSP14}, Kozinski~\cite{DBLP:conf/cvpr/KozinskiGZOM15}, Cohen~\cite{DBLP:conf/3dim/CohenOLP17} and DeepFacade~\cite{liu2020deepfacade}. Note that only DeepFacade~\cite{liu2020deepfacade} reported IOU scores on this dataset.}
	\label{tab:chap5-artdeco}
\end{table*}

\begin{table*}[htbp]
	\centering
	\begin{tabular}{|c|c|c|c|c|c|c|c|c|c|}
		\hline
		Method & Window & Wall & Balcony & Door & Roof & Sky & Shop & Chimney & mIOU\\
		\hline
		Before &80.3 &89.8 &85.2 &63.1 &75.6 &88.4 &83.8 &68.5 & 79.3\\
		\hline
		After &82.5 &91.6 &87.2 &70.6 &76.2 &88.4 &84.3 &68.5 & 81.2\\
		\hline
	\end{tabular}
	\caption{Ablation study on the ECP dataset(\%). Before means before the translational symmetry based refinement, the same goes for After.}
	\label{tab:ecp-ablation}
\end{table*}

\begin{figure*}[htbp]
	\centering
	\subfloat{\includegraphics[height=0.19\textheight]{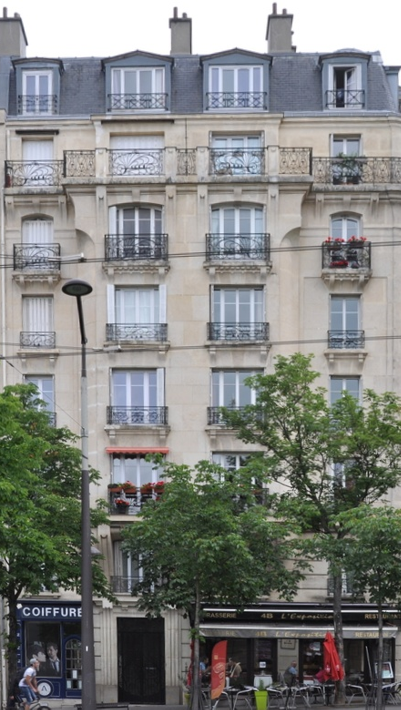}}
	\hspace{.1em}
	\subfloat{\includegraphics[height=0.19\textheight]{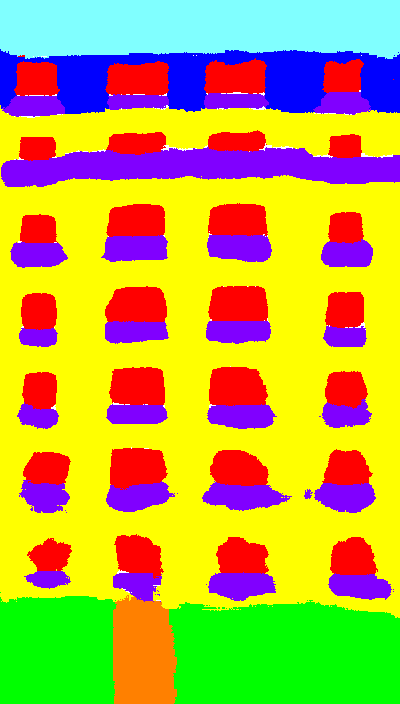}}
	\hspace{.1em}
	\subfloat{\includegraphics[height=0.19\textheight]{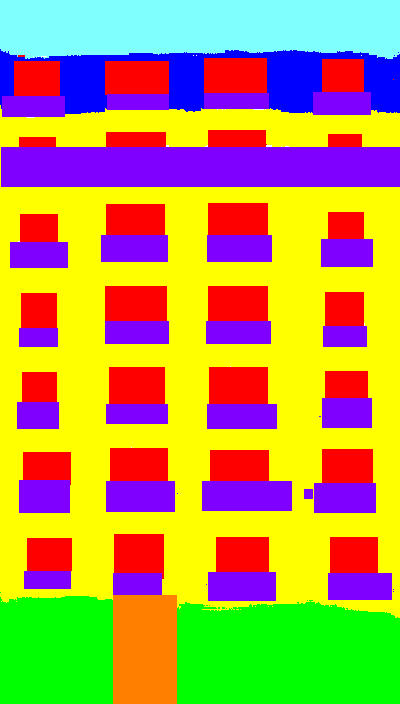}}
	\hspace{.1em}
	\subfloat{\includegraphics[height=0.19\textheight]{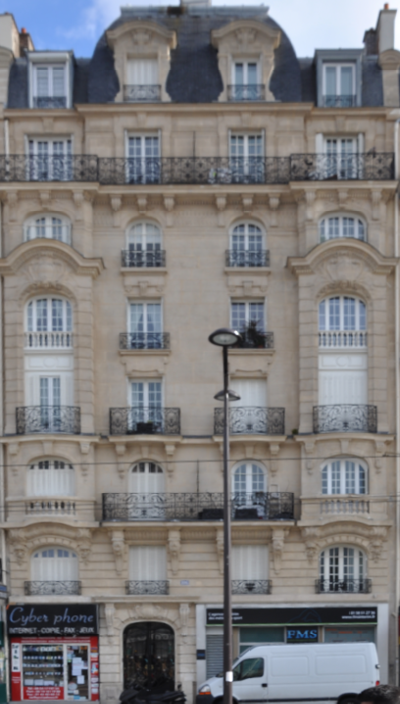}}
	\hspace{.1em}
	\subfloat{\includegraphics[height=0.19\textheight]{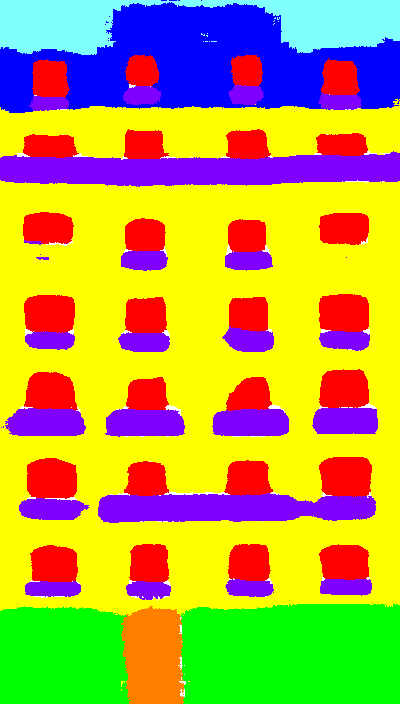}}
	\hspace{.1em}
	\subfloat{\includegraphics[height=0.19\textheight]{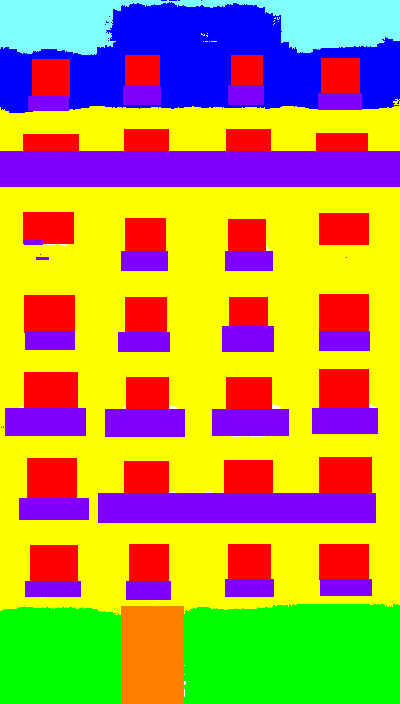}}
	\\
	\subfloat{\includegraphics[height=0.19\textheight]{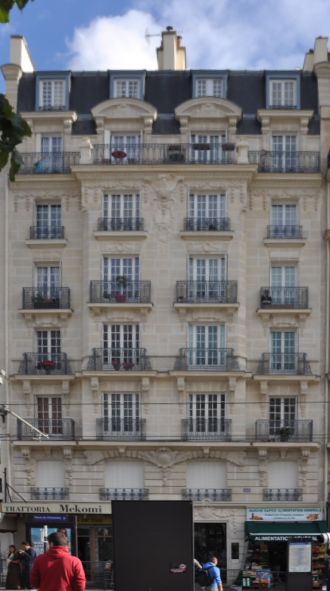}}
	\hspace{.1em}
	\subfloat{\includegraphics[height=0.19\textheight]{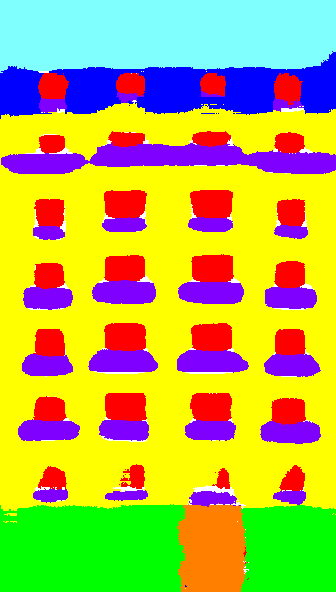}}
	\hspace{.1em}
	\subfloat{\includegraphics[height=0.19\textheight]{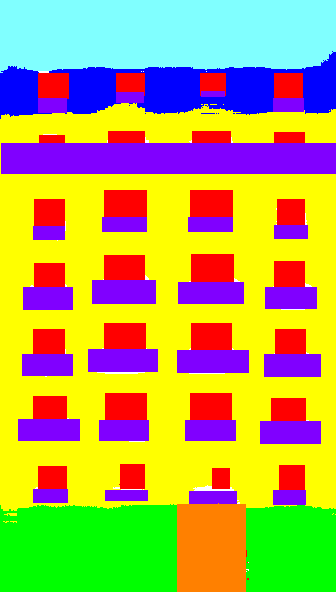}}
	\hspace{.1em}
	\subfloat{\includegraphics[height=0.19\textheight]{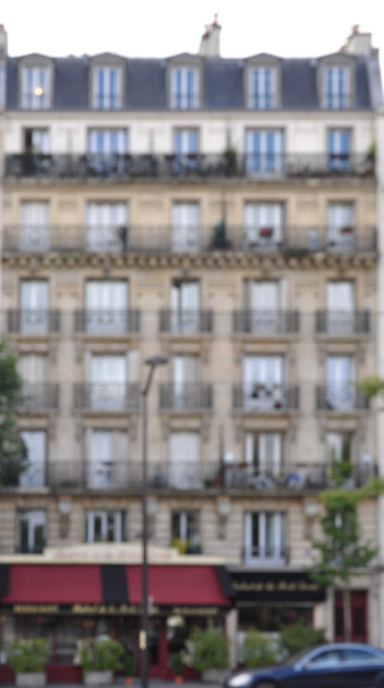}}
	\hspace{.1em}
	\subfloat{\includegraphics[height=0.19\textheight]{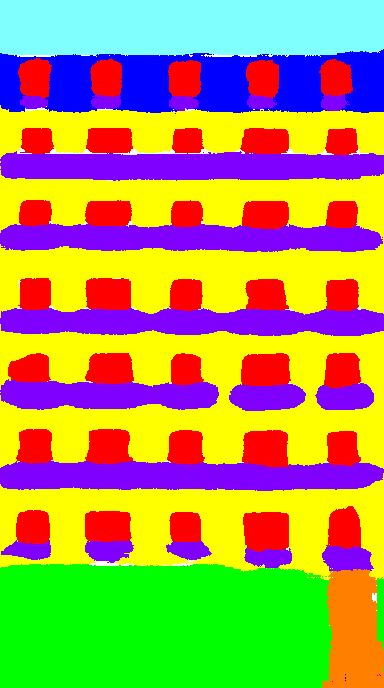}}
	\hspace{.1em}
	\subfloat{\includegraphics[height=0.19\textheight]{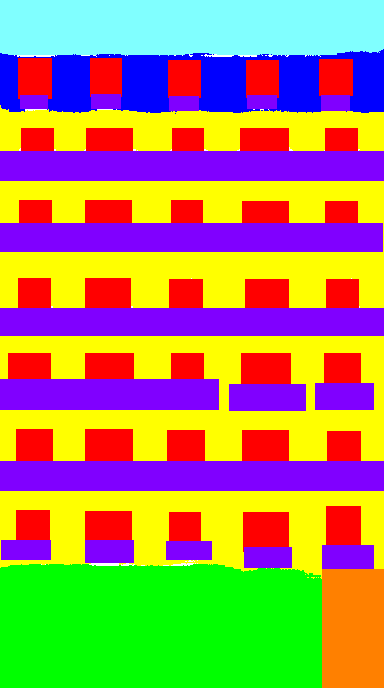}}
	\caption{Visual results on the ArtDeco dataset. From left to right in each group, we show the input image, visual results predicted by the neural network, segmentation map after the translational symmetry refinement, respectively.}
	\label{fig:chap5-artdeco-visual}
\end{figure*}
\subsubsection{Results on ArtDeco dataset}  Table~\ref{tab:chap5-artdeco} gives the quantitative results on the ArtDeco dataset. We compare with four recent methods, including Cohen~\cite{DBLP:conf/cvpr/CohenSP14}, Kozinski~\cite{DBLP:conf/cvpr/KozinskiGZOM15}, Cohen~\cite{DBLP:conf/3dim/CohenOLP17} and DeepFacade~\cite{liu2020deepfacade}. It can be seen that our approach achieves the best pixel accuracy of $93.5\%$. In particular, our method clearly performs the best on window and balcony. As for IOU scores, our proposed approach outperforms  DeepFacade~\cite{liu2020deepfacade} over 4 percent. 

The above quantitative results demonstrate the efficacy on facade parsing of our presented method.

\begin{figure*}[htbp]
	\centering
	\subfloat{\includegraphics[width=0.153\textwidth,height=0.18\textheight]{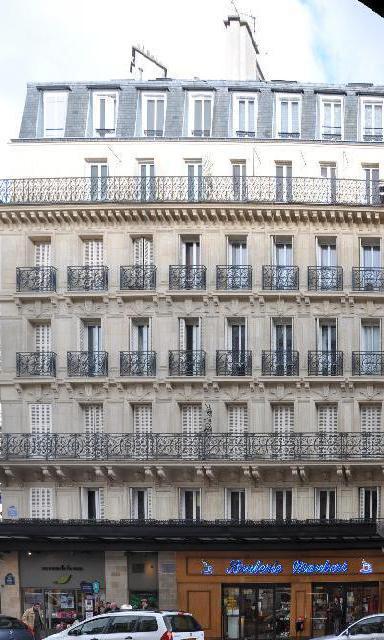}}
	\hspace{.1em}
	\subfloat{\includegraphics[width=0.153\textwidth,height=0.18\textheight]{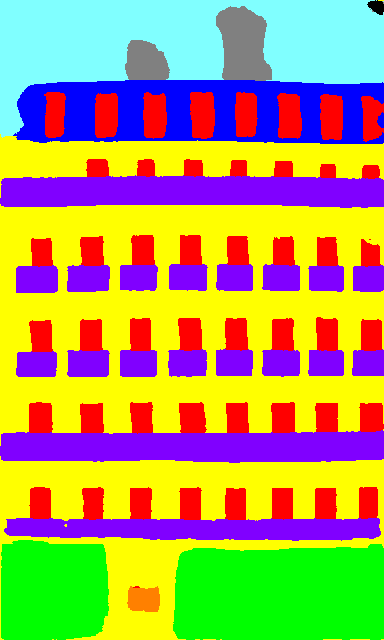}}
	\hspace{.1em}
	\subfloat{\includegraphics[width=0.153\textwidth,height=0.18\textheight]{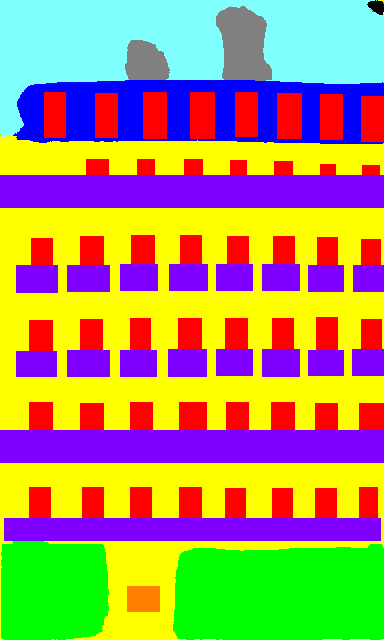}}
	\hspace{.1em}
	\subfloat{\includegraphics[width=0.153\textwidth,height=0.18\textheight]{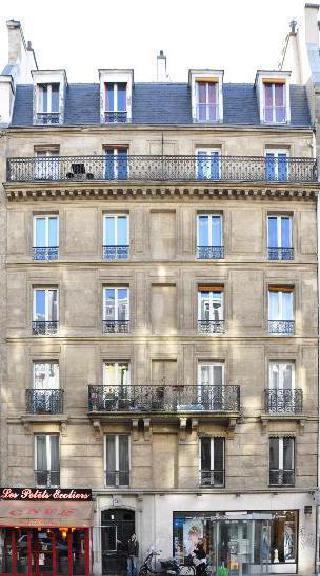}}
	\hspace{.1em}
	\subfloat{\includegraphics[width=0.153\textwidth,height=0.18\textheight]{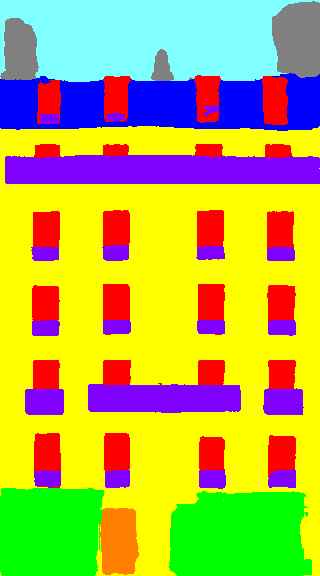}}
	\hspace{.1em}
	\subfloat{\includegraphics[width=0.153\textwidth,height=0.18\textheight]{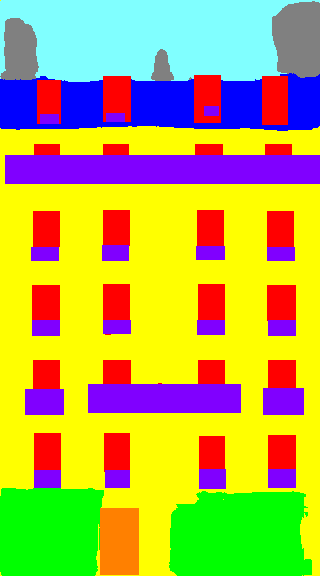}}
	\hspace{.1em}
	\\
	\subfloat{\includegraphics[width=0.153\textwidth,height=0.18\textheight]{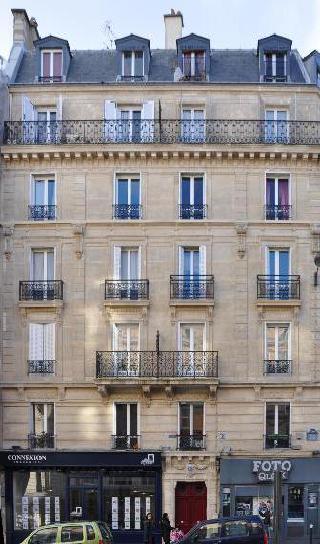}}
	\hspace{.1em}
	\subfloat{\includegraphics[width=0.153\textwidth,height=0.18\textheight]{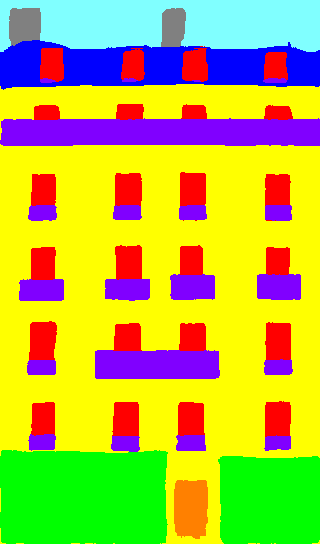}}
	\hspace{.1em}
	\subfloat{\includegraphics[width=0.153\textwidth,height=0.18\textheight]{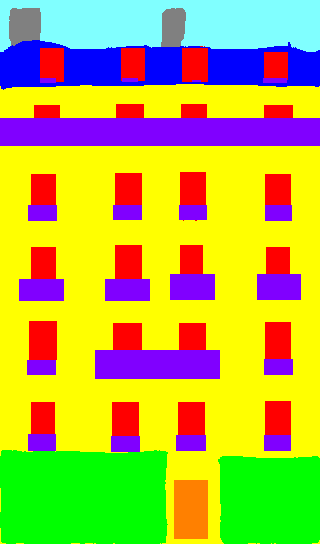}}
	\hspace{.1em}
	\subfloat{\includegraphics[width=0.153\textwidth,height=0.18\textheight]{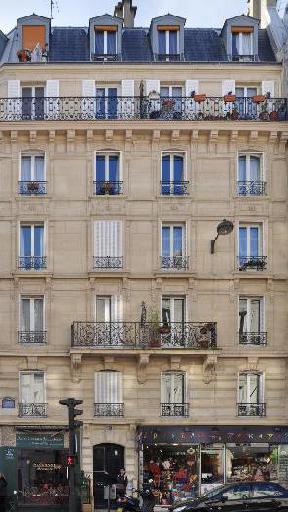}}
	\hspace{.1em} 
	\subfloat{\includegraphics[width=0.153\textwidth,height=0.18\textheight]{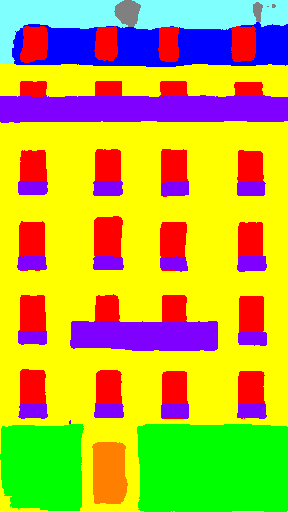}}
	\hspace{.1em}
	\subfloat{\includegraphics[width=0.153\textwidth,height=0.18\textheight]{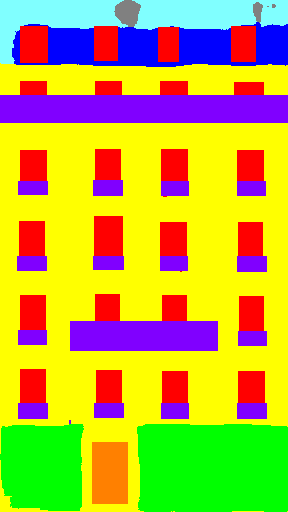}}
	\hspace{.1em}
	\\
	\subfloat{\includegraphics[height=0.17\textheight]{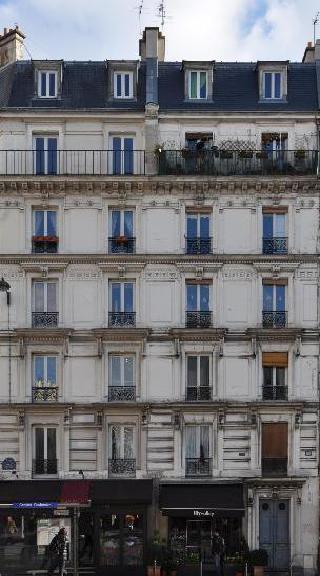}}
	\hspace{.1em}
	\subfloat{\includegraphics[height=0.17\textheight]{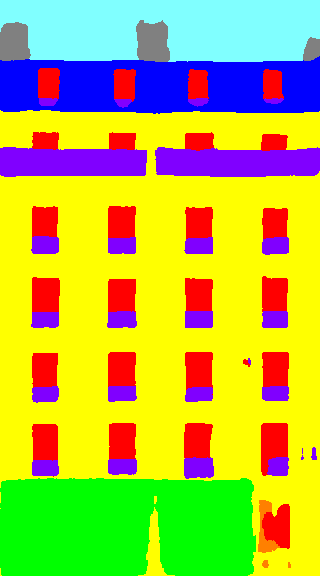}}
	\hspace{.1em}
	\subfloat{\includegraphics[height=0.17\textheight]{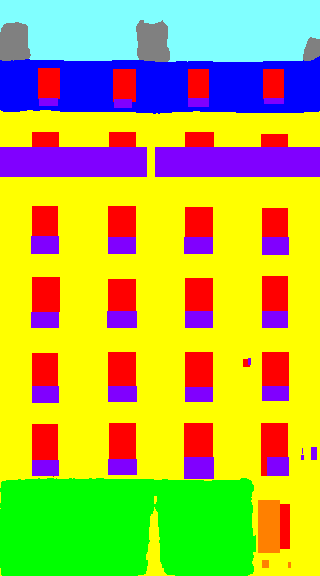}}
	\hspace{.1em}
	\subfloat{\includegraphics[width=0.179\textwidth,height=0.17\textheight]{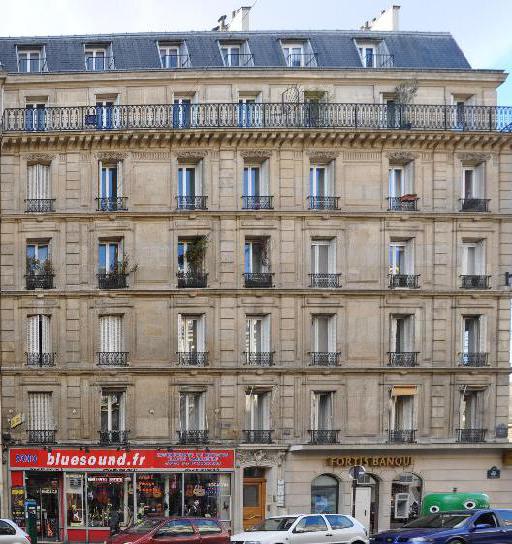}}
	\hspace{.1em}
	\subfloat{\includegraphics[width=0.179\textwidth,height=0.17\textheight]{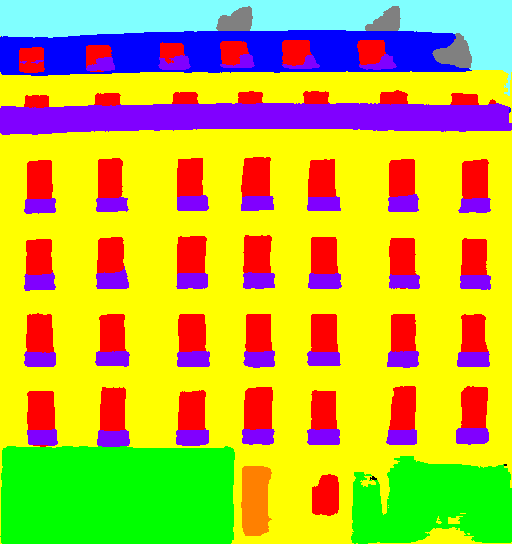}}
	\hspace{.1em}
	\subfloat{\includegraphics[width=0.179\textwidth,height=0.17\textheight]{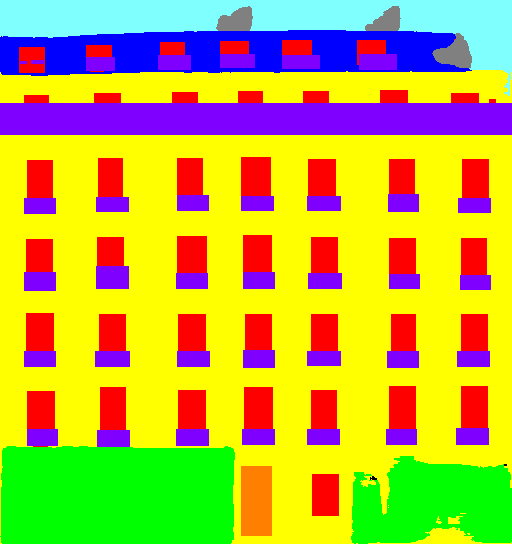}}
	\caption{Visual results on the ECP dataset. From left to right are the input image, the result before refinement and the result after refinement.}
	\label{fig:translational}
\end{figure*}

\subsection{Ablation Study}
The proposed neural network architecture consists of several components. Moreover, we are interested in examining the efficacy of translational symmetry refinement. To this end, we conduct a series of ablation study to analyze how each part will affect the final results.

Our network consists of two branches, namely semantic segmentation and object detection. It is interested to study how each head of our network affects the accuracy of final results. We conducted the ablation study on the RueMonge dataset. Specifically, we train our network with three different settings, semantic segmentation head only, detection head only and both heads simultaneously. 


Table~\ref{tab:network-ablation} shows the IOUs scores for each setting. The `Segmentation' column is the result obtained by training with the semantic segmentation head only. The `Detection' column is the result training with only anchor-free detection results. The `Fusion' column is the full network training with both heads. Since the detectors are only trained with the classes of window, balcony and door, we only report the accuracy for these three classes in the `Detection' column.

\begin{table}[htbp]
	\centering
	\begin{tabular}{|c|c|c|c|}
		\hline
		Class & Segmentation & Detection & Fusion\\
		\hline
		Window & 60.58 & 60.71 & 60.94\\
		\hline
		Wall & 78.83 & - &79.63\\
		\hline
		Balcony & 72.09& 72.31 &72.66\\
		\hline
		Door & 26.69& 27.15 &27.19\\
		\hline
		Roof & 65.51& - &66.08\\
		\hline
		Sky & 86.68& - &87.16\\
		\hline
		Shop & 63.32& - &63.81\\
		\hline
		mIOU & 64.81& - &65.35\\
		\hline
	\end{tabular}
	\caption{Ablation study of neural network(\%). We have tested our method in three different settings on the RueMonge dataset, including semantic segmentation head only, detection head only and both heads simultaneously.}
	
	\vspace{-0.2in}
	\label{tab:network-ablation}
\end{table}

It can be seen that semantic segmentation alone can not yield the optimal results on the RueMonge dataset. The IOUs of the window, door and balcony are slightly lower than the results of training with the detection head. When training them together, the fusion method achieves the best accuracy. This demonstrates the efficacy of our method, and the anchor-free detection head indeed improves the accuracy.

To study the effectiveness of translational symmetry-based refinement, we conduct ablation study on the ECP dataset with two different settings. One directly outputs of the neural network while another refines the inference results. Table~\ref{tab:ecp-ablation} shows that the refinement module generally increase the IOU score at around 2 percent. This demonstrates the effectiveness of our proposed method.

\subsection{Qualitative Evaluation}

\begin{figure*}[htbp]
	\centering
	\includegraphics[width=1.0\textwidth]{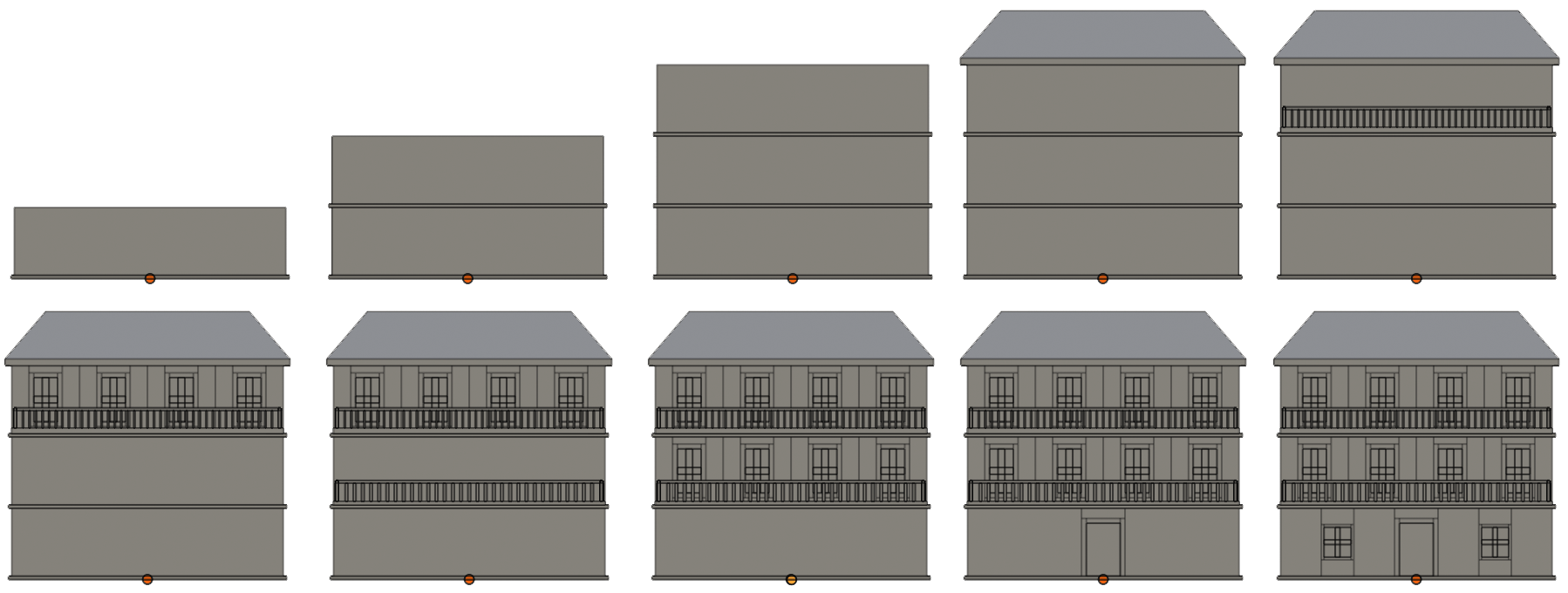}
	\caption{Mesh generation for a 3d building model. We start from the silhouette of each floor, and then gradually add its elements.}
	\label{fig:mesh-process}
\end{figure*}

\begin{figure*}[t]
	\centering
	\subfloat{
		\includegraphics[width=0.135\textwidth]{ecp_monge_32.jpg} 
	}
	\subfloat{
		\includegraphics[width=0.135\textwidth]{ecp_monge_101.jpg}
	}
	\subfloat{
		\includegraphics[width=0.234\textwidth]{ecp_monge_106.jpg}
	}
	\subfloat{
		\includegraphics[width=0.135\textwidth]{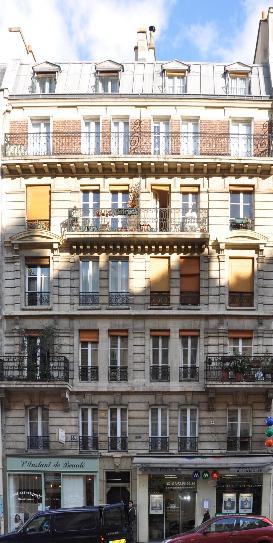}
	}
	\subfloat{
		\includegraphics[width=0.135\textwidth]{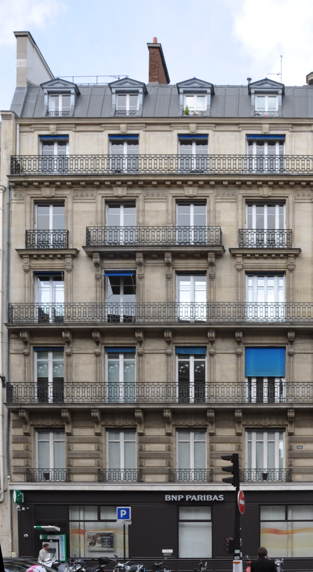}
	}
	\\
	\subfloat{
		\includegraphics[width=0.135\textwidth]{ecp_better_monge_32.png} 
	}
	\subfloat{
		\includegraphics[width=0.135\textwidth]{ecp_better_monge_101.png}
	}
	\subfloat{
		\includegraphics[width=0.234\textwidth]{ecp_better_monge_106.png}
	}
	\subfloat{
		\includegraphics[width=0.135\textwidth]{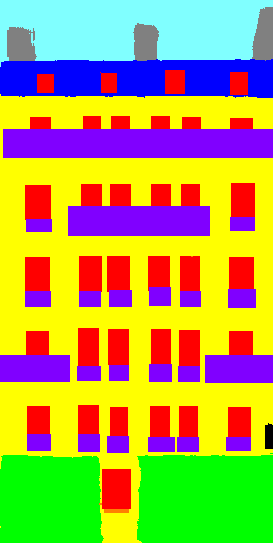}
	}
	\subfloat{
		\includegraphics[width=0.135\textwidth]{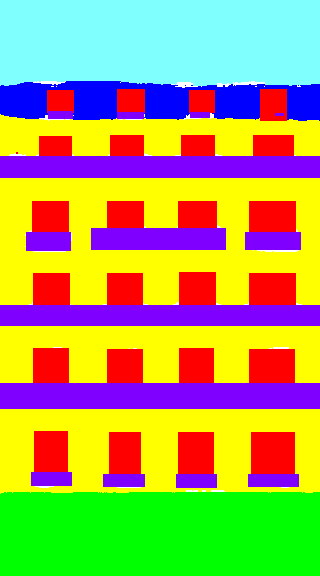}
	}
	\\
	\subfloat{
		\includegraphics[width=0.85\textwidth]{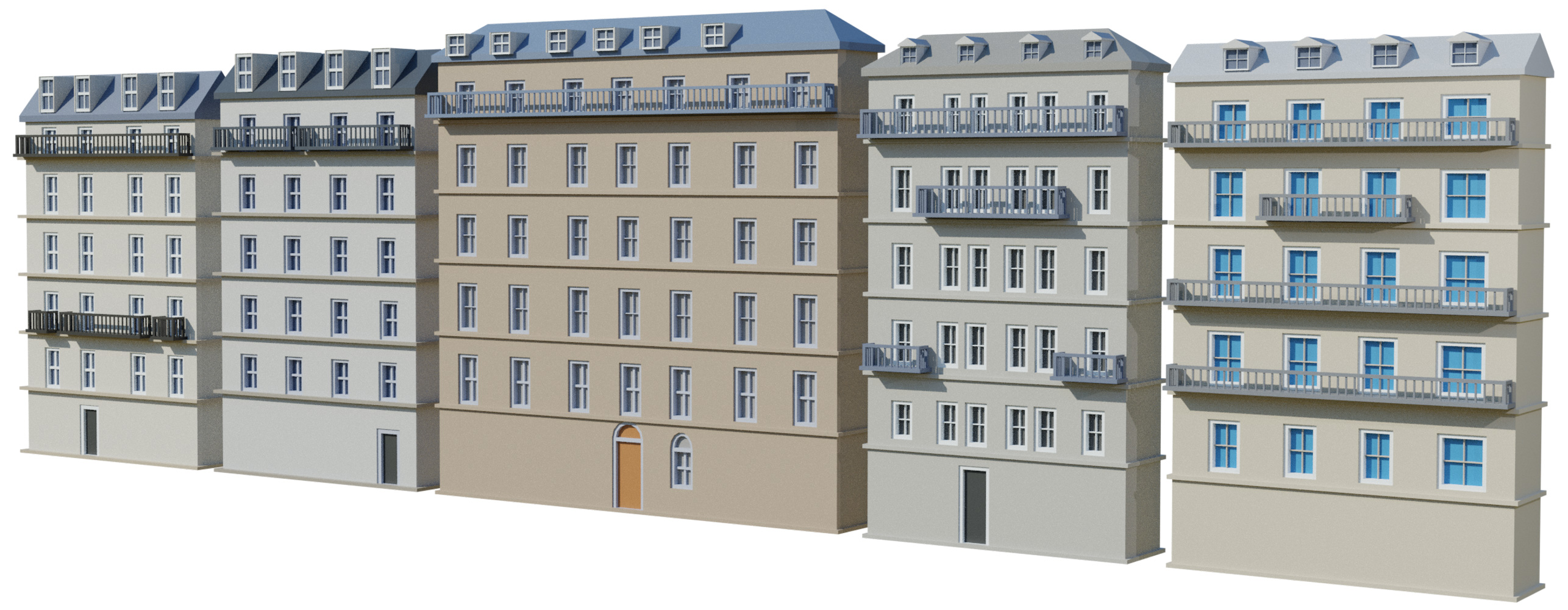}
	}
	\caption{3D building reconstruction results. From top to bottom are the input images, facade parsing results and reconstructed 3D models, respectively.}
	\label{fig:chap6-visual}
\end{figure*}

In the following, we show the qualitative results of our method on different testbeds.

Fig.~\ref{fig:chap5-artdeco-visual} shows the visual results of our presented method on the ArtDeco dataset, which consists of occlusion caused by vegetation. As in the upper-left group, there are trees in front of buildings. It can be observed that our method has predicted the rough position of windows behind the trees. Although the guess is not perfectly accurate, it is an acceptable prediction. After the translational symmetry refinement, the overall visual result has improved significantly. Occlusions may affect the translational symmetric score, and the refinement process tends to keep the original size of the objects. In case of no occlusions, the sizes are more evenly distributed than the occluded cases.

Fig.~\ref{fig:translational} shows the visual results on the ECP dataset. It can be seen that the sizes of windows, balconies are more unified than before the refinement, which is more visual pleasant and closer to the ground truth. As the facades in this dataset are more translational symmetric than the ArtDeco, we can obtain better results.

The above experimental results demonstrate that our method not only obtains the high parsing accuracy, but also show the promising visual predictions.

\subsection{Computational Time Evaluation}
\begin{table}[htbp]
  \centering
  \begin{tabular}{|c|c|c|c|c|}
    \hline
    Method & ~\cite{DBLP:conf/cvpr/MartinovicKRG15} 2D & ~\cite{DBLP:conf/cvpr/MartinovicKRG15} 2D+3D & DeepFacade~\cite{liu2020deepfacade} & Ours\\
    \hline
    Test Time & 379min & 470min & 30s & 17s\\
    \hline
  \end{tabular}
  \caption{Testing time cost comparison with state-of-the-art methods. The time cost in each entry is the total time cost to predict the whole test set.}
  \label{tab:time-cost}
\end{table}

Table~\ref{tab:time-cost} shows the testing time for different methods, where each entry is the total time of predicting all images in the testing dataset. Martinovic et al.~\cite{DBLP:conf/cvpr/MartinovicKRG15} require the highest computational cost, since it involves the complex CRF inference stage. Their 2D method costs 379 minutes for testing, and the 2D + 3D approach needs 470 minutes.  Deep learning-based methods are relatively fast in inference. DeepFacade~\cite{liu2020deepfacade} processes the whole testing set within 30s while our presented method only needs 17s. This is because our anchor-free framework just computes the feed-forward network pass once rather than twice for the two stage method in DeepFacade.

As for the training time, DeepFacade~\cite{liu2020deepfacade} has to train two different kinds of networks while our proposed approach only needs to train a single network. This means that our network costs half the time of DeepFacade during training if both methods use the same backbone network. In practice, the training of our presented neural network can be done within two hours due to the small scale of the dataset.

The above experiment indicates that our proposed approach is highly efficient. 



\section{3D Building Reconstruction}
Once the facade is parsed into semantic grammars, we can reconstruct their 3D model by procedural modeling. In this paper, Blender~\cite{blender} with Python API is employed in our implementation.

The first step is to reconstruct the geometric meshes of the building facade. We adopt a modular way to achieve this goal. Each facade is divided into the modular components that can be assembled. As shown in Fig.~\ref{fig:mesh-process}, we start from building each floor of the facades, and then gradually add windows and other elements to it. Each object is a four-tuple $obj=(x,y,w,h)$, where $(x,y)$ is the center location and $(w,h)$ is the size. Note that we have a predefined shape template for each object. In this process, we actually scale the template to target size, and then place the scaled template into the target location.

After generating the meshes of a facade, we need to estimate the materials of 3D building model. Therefore, we have to predict the material from 2D images. Since our goal is to perform the semantic-level reconstruction, we do not seek to restore the building facade perfectly without any error.  In this paper, we focus on the overall semantic result instead. In particular, we employ the average color as the material for wall, balcony, door, frames and roof. For window glasses, we use a kind of blue color that is close to the original image. After obtaining the materials, we finish the modeling process and render the result.

\begin{figure*}[hbtp]
	\centering
	\subfloat[Input]{
		\includegraphics[width=0.15\textwidth,height=0.2\textheight]{ecp_monge_101.jpg} 
	}
	\subfloat[DeepFacade\cite{DBLP:conf/ijcai/LiuZZH17}]{
		\includegraphics[width=0.15\textwidth,height=0.2\textheight]{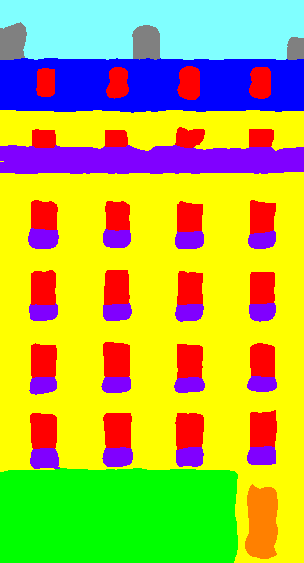} 
	}
	\subfloat[Ours]{
		\includegraphics[width=0.15\textwidth,height=0.2\textheight]{ecp_better_monge_101.png} 
	}
	\subfloat[DeepFacade\cite{DBLP:conf/ijcai/LiuZZH17}]{
		\includegraphics[width=0.15\textwidth,height=0.18\textheight]{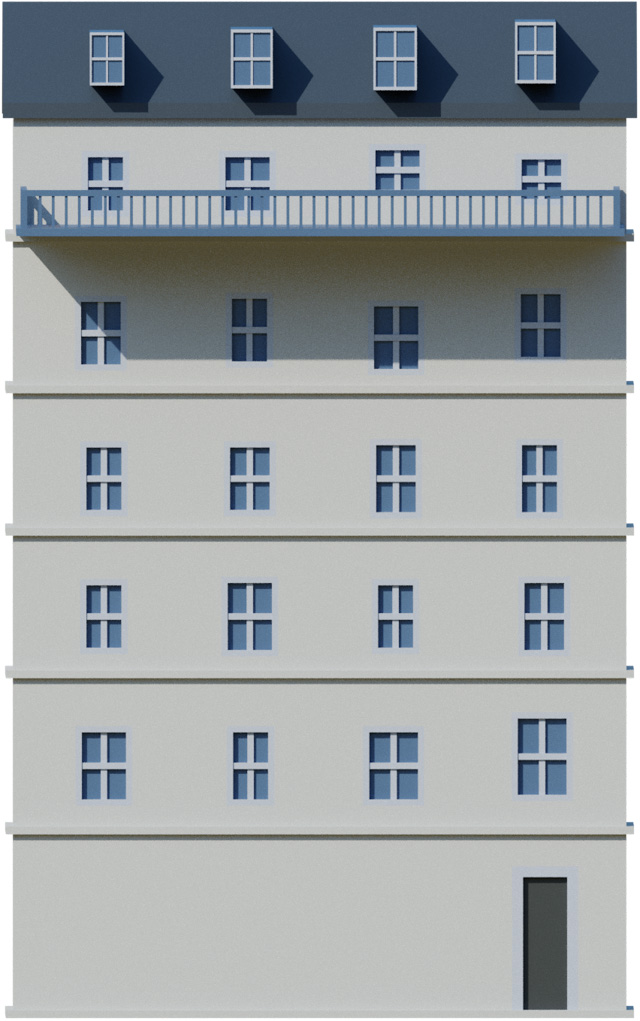} 
	}
	\subfloat[Ours]{
		\includegraphics[width=0.17\textwidth,height=0.182\textheight]{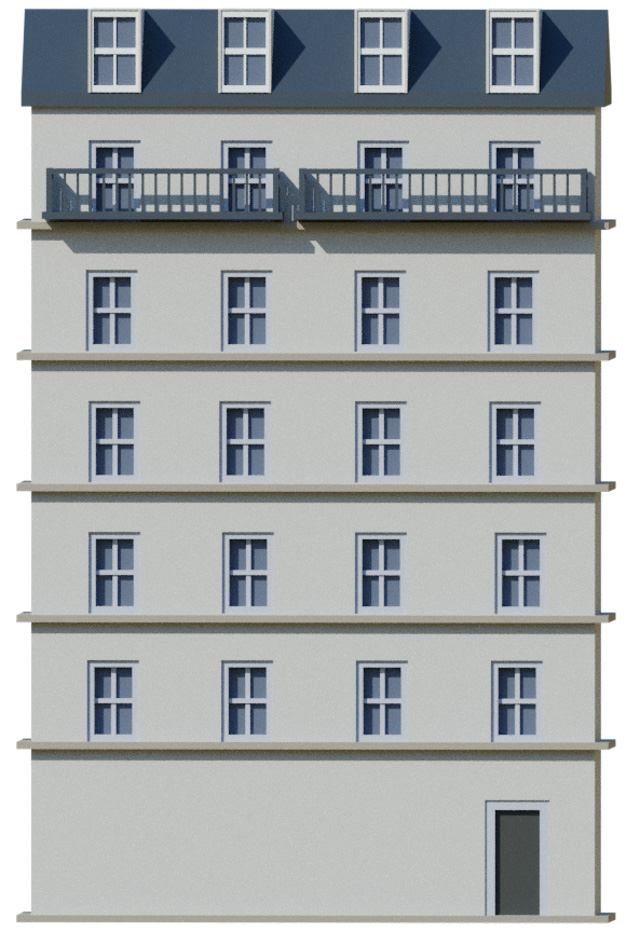} 
	}
	\caption{Building reconstruction results comparison for the different facade parsing methods.}
	\label{fig:chap6-compare35}
\end{figure*}

Fig.~\ref{fig:intro-demo} illustrates the pipeline of our proposed approach for a single building. Fig.~\ref{fig:chap6-visual} shows more reconstruction results and their street view. For aesthetic reasons, we omit balconies that are too small during rendering and only render those with large size. This makes the model look more promising. Fig.~\ref{fig:chap6-compare35} compares the reconstruction results using different parsing methods. We compare our result with DeepFacade~\cite{DBLP:conf/ijcai/LiuZZH17}. It can been seen that the better parsing results will yield a more unified and visually satisfied 3D model.

\section{Conclusions}
In this paper, we have proposed an anchor-free neural network fusing semantic segmentation and facade object detection for 3D building reconstruction. Moreover, we have introduced a novel translational symmetry-based refinement module to refine the predicted segmentation map. Furthermore, we have presented a procedural modeling pipeline to reconstruct the 3D building from the facade parsing results. Finally, we have performed both quantitative and qualitative experiments on three popular facade parsing datasets, whose promising results demonstrate the efficacy of our proposed translational symmetry-aware approach. 

For future work, we may explore the techniques to directly regress the parametric procedural models from the image rather than resorting to the pixel-labeled segmentation map. This will further improve the parsing accuracy and simplify the 3D building reconstruction pipeline.



\ifCLASSOPTIONcaptionsoff
  \newpage
\fi

\bibliographystyle{IEEEtran}
\bibliography{facade3d}

\begin{thebibliography}{10}
\providecommand{\url}[1]{#1}
\csname url@samestyle\endcsname
\providecommand{\newblock}{\relax}
\providecommand{\bibinfo}[2]{#2}
\providecommand{\BIBentrySTDinterwordspacing}{\spaceskip=0pt\relax}
\providecommand{\BIBentryALTinterwordstretchfactor}{4}
\providecommand{\BIBentryALTinterwordspacing}{\spaceskip=\fontdimen2\font plus
\BIBentryALTinterwordstretchfactor\fontdimen3\font minus
  \fontdimen4\font\relax}
\providecommand{\BIBforeignlanguage}[2]{{%
\expandafter\ifx\csname l@#1\endcsname\relax
\typeout{** WARNING: IEEEtran.bst: No hyphenation pattern has been}%
\typeout{** loaded for the language `#1'. Using the pattern for}%
\typeout{** the default language instead.}%
\else
\language=\csname l@#1\endcsname
\fi
#2}}
\providecommand{\BIBdecl}{\relax}
\BIBdecl

\bibitem{teboul2010segmentation}
O.~Teboul, L.~Simon, P.~Koutsourakis, and N.~Paragios, ``Segmentation of
  building facades using procedural shape priors,'' in \emph{Computer Vision
  and Pattern Recognition (CVPR), 2010 IEEE Conference on}.\hskip 1em plus
  0.5em minus 0.4em\relax IEEE, 2010, pp. 3105--3112.

\bibitem{martinovic2012three}
A.~Martinovi{\'c}, M.~Mathias, J.~Weissenberg, and L.~Van~Gool, ``A
  three-layered approach to facade parsing,'' in \emph{European conference on
  computer vision}.\hskip 1em plus 0.5em minus 0.4em\relax Springer, 2012, pp.
  416--429.

\bibitem{liu2020deepfacade}
H.~Liu, Y.~Xu, J.~Zhang, J.~Zhu, Y.~Li, and C.~S. Hoi, ``Deepfacade: A deep
  learning approach to facade parsing with symmetric loss,'' \emph{IEEE
  Transactions on Multimedia}, 2020.

\bibitem{DBLP:journals/cgf/NishidaBA18}
G.~Nishida, A.~Bousseau, and D.~G. Aliaga, ``Procedural modeling of a building
  from a single image,'' \emph{Comput. Graph. Forum}, vol.~37, no.~2, pp.
  415--429, 2018.

\bibitem{DBLP:journals/corr/abs-1904-07850}
X.~Zhou, D.~Wang, and P.~Kr{\"{a}}henb{\"{u}}hl, ``Objects as points,''
  \emph{CoRR}, vol. abs/1904.07850, 2019.

\bibitem{DBLP:conf/cvpr/ZhouZK19}
X.~Zhou, J.~Zhuo, and P.~Kr{\"{a}}henb{\"{u}}hl, ``Bottom-up object detection
  by grouping extreme and center points,'' in \emph{{CVPR}}.\hskip 1em plus
  0.5em minus 0.4em\relax Computer Vision Foundation / {IEEE}, 2019, pp.
  850--859.

\bibitem{DBLP:conf/cvpr/RedmonDGF16}
J.~Redmon, S.~K. Divvala, R.~B. Girshick, and A.~Farhadi, ``You only look once:
  Unified, real-time object detection,'' in \emph{{CVPR}}, 2016, pp. 779--788.

\bibitem{blender}
\BIBentryALTinterwordspacing
B.~O. Community, \emph{Blender - a 3D modelling and rendering package}, Blender
  Foundation, Stichting Blender Foundation, Amsterdam, 2018. [Online].
  Available: \url{http://www.blender.org}
\BIBentrySTDinterwordspacing

\bibitem{DBLP:journals/tog/MullerWHUG06}
P.~M{\"{u}}ller, P.~Wonka, S.~Haegler, A.~Ulmer, and L.~V. Gool, ``Procedural
  modeling of buildings,'' \emph{{ACM} Trans. Graph.}, vol.~25, no.~3, pp.
  614--623, 2006.

\bibitem{DBLP:conf/nips/KrizhevskySH12}
A.~Krizhevsky, I.~Sutskever, and G.~E. Hinton, ``Imagenet classification with
  deep convolutional neural networks,'' in \emph{{NIPS}}, 2012, pp. 1106--1114.

\bibitem{DBLP:journals/tog/MullerZWG07}
P.~M{\"{u}}ller, G.~Zeng, P.~Wonka, and L.~V. Gool, ``Image-based procedural
  modeling of facades,'' \emph{{ACM} Trans. Graph.}, vol.~26, no.~3, p.~85,
  2007.

\bibitem{DBLP:conf/cvpr/MartinovicKRG15}
A.~Martinovic, J.~Knopp, H.~Riemenschneider, and L.~V. Gool, ``3d all the way:
  Semantic segmentation of urban scenes from start to end in 3d,'' in
  \emph{{CVPR}}, 2015, pp. 4456--4465.

\bibitem{DBLP:conf/cvpr/KozinskiGZOM15}
M.~Kozinski, R.~Gadde, S.~Zagoruyko, G.~Obozinski, and R.~Marlet, ``A {MRF}
  shape prior for facade parsing with occlusions,'' in \emph{{CVPR}}, 2015, pp.
  2820--2828.

\bibitem{DBLP:conf/cvpr/ZhaoFXZZQ10}
P.~Zhao, T.~Fang, J.~Xiao, H.~Zhang, Q.~Zhao, and L.~Quan, ``Rectilinear
  parsing of architecture in urban environment,'' in \emph{{CVPR}}, 2010, pp.
  342--349.

\bibitem{DBLP:conf/dagm/WendelDB10}
A.~Wendel, M.~Donoser, and H.~Bischof, ``Unsupervised facade segmentation using
  repetitive patterns,'' in \emph{DAGM-Symposium}, ser. Lecture Notes in
  Computer Science, vol. 6376.\hskip 1em plus 0.5em minus 0.4em\relax Springer,
  2010, pp. 51--60.

\bibitem{DBLP:conf/3dim/ReckyWL11}
M.~Recky, A.~Wendel, and F.~Leberl, ``Fa{\c{c}}ade segmentation in a multi-view
  scenario,'' in \emph{3DIMPVT}, 2011, pp. 358--365.

\bibitem{DBLP:conf/iccv/KoutsourakisSTTP09}
P.~Koutsourakis, L.~Simon, O.~Teboul, G.~Tziritas, and N.~Paragios, ``Single
  view reconstruction using shape grammars for urban environments,'' in
  \emph{{ICCV}}, 2009, pp. 1795--1802.

\bibitem{ripperda2006reconstruction}
N.~Ripperda and C.~Brenner, ``Reconstruction of fa{\c{c}}ade structures using a
  formal grammar and rjmcmc,'' in \emph{Joint Pattern Recognition
  Symposium}.\hskip 1em plus 0.5em minus 0.4em\relax Springer, 2006, pp.
  750--759.

\bibitem{teboul2011shape}
O.~Teboul, I.~Kokkinos, L.~Simon, P.~Koutsourakis, and N.~Paragios, ``Shape
  grammar parsing via reinforcement learning,'' in \emph{Computer Vision and
  Pattern Recognition (CVPR), 2011 IEEE Conference on}.\hskip 1em plus 0.5em
  minus 0.4em\relax IEEE, 2011, pp. 2273--2280.

\bibitem{DBLP:journals/ijcv/MathiasMG16}
\BIBentryALTinterwordspacing
M.~Mathias, A.~Martinovic, and L.~V. Gool, ``{ATLAS:} {A} three-layered
  approach to facade parsing,'' \emph{International Journal of Computer
  Vision}, vol. 118, no.~1, pp. 22--48, 2016. [Online]. Available:
  \url{http://dx.doi.org/10.1007/s11263-015-0868-z}
\BIBentrySTDinterwordspacing

\bibitem{Cohen_2014_CVPR}
A.~Cohen, A.~G. Schwing, and M.~Pollefeys, ``Efficient structured parsing of
  facades using dynamic programming,'' June 2014.

\bibitem{DBLP:conf/ijcai/LiuZZH17}
H.~Liu, J.~Zhang, J.~Zhu, and S.~C.~H. Hoi, ``Deepfacade: {A} deep learning
  approach to facade parsing,'' in \emph{{IJCAI}}.\hskip 1em plus 0.5em minus
  0.4em\relax ijcai.org, 2017, pp. 2301--2307.

\bibitem{schmitz2016convolutional}
M.~Schmitz and H.~Mayer, ``A convolutional network for semantic facade
  segmentation and interpretation.'' \emph{ISPRS-International Archives of the
  Photogrammetry, Remote Sensing and Spatial Information Sciences}, pp.
  709--715, 2016.

\bibitem{DBLP:journals/corr/abs-1805-08634}
J.~Femiani, W.~R. Para, N.~J. Mitra, and P.~Wonka, ``Facade segmentation in the
  wild,'' \emph{CoRR}, vol. abs/1805.08634, 2018.

\bibitem{long2015fully}
J.~Long, E.~Shelhamer, and T.~Darrell, ``Fully convolutional networks for
  semantic segmentation,'' in \emph{Proceedings of the IEEE Conference on
  Computer Vision and Pattern Recognition}, 2015, pp. 3431--3440.

\bibitem{DBLP:journals/corr/ChenPKMY14}
L.~Chen, G.~Papandreou, I.~Kokkinos, K.~Murphy, and A.~L. Yuille, ``Semantic
  image segmentation with deep convolutional nets and fully connected crfs,''
  in \emph{{ICLR} (Poster)}, 2015.

\bibitem{DBLP:journals/corr/ChenPSA17}
L.~Chen, G.~Papandreou, F.~Schroff, and H.~Adam, ``Rethinking atrous
  convolution for semantic image segmentation,'' \emph{CoRR}, vol.
  abs/1706.05587, 2017.

\bibitem{DBLP:journals/pami/ChenPKMY18}
L.~Chen, G.~Papandreou, I.~Kokkinos, K.~Murphy, and A.~L. Yuille, ``Deeplab:
  Semantic image segmentation with deep convolutional nets, atrous convolution,
  and fully connected crfs,'' \emph{{IEEE} Trans. Pattern Anal. Mach. Intell.},
  vol.~40, no.~4, pp. 834--848, 2018.

\bibitem{DBLP:journals/pami/BadrinarayananK17}
V.~Badrinarayanan, A.~Kendall, and R.~Cipolla, ``Segnet: {A} deep convolutional
  encoder-decoder architecture for image segmentation,'' \emph{{IEEE} Trans.
  Pattern Anal. Mach. Intell.}, vol.~39, no.~12, pp. 2481--2495, 2017.

\bibitem{DBLP:conf/cvpr/LinDGHHB17}
T.~Lin, P.~Doll{\'{a}}r, R.~B. Girshick, K.~He, B.~Hariharan, and S.~J.
  Belongie, ``Feature pyramid networks for object detection,'' in
  \emph{{CVPR}}, 2017, pp. 936--944.

\bibitem{DBLP:conf/cvpr/ZhaoSQWJ17}
H.~Zhao, J.~Shi, X.~Qi, X.~Wang, and J.~Jia, ``Pyramid scene parsing network,''
  in \emph{{CVPR}}, 2017, pp. 6230--6239.

\bibitem{DBLP:conf/miccai/RonnebergerFB15}
O.~Ronneberger, P.~Fischer, and T.~Brox, ``U-net: Convolutional networks for
  biomedical image segmentation,'' in \emph{{MICCAI} {(3)}}, ser. Lecture Notes
  in Computer Science, vol. 9351.\hskip 1em plus 0.5em minus 0.4em\relax
  Springer, 2015, pp. 234--241.

\bibitem{DBLP:journals/tmm/AbdulnabiSZCW18}
A.~H. Abdulnabi, B.~Shuai, Z.~Zuo, L.~Chau, and G.~Wang, ``Multimodal recurrent
  neural networks with information transfer layers for indoor scene labeling,''
  \emph{{IEEE} Trans. Multimedia}, vol.~20, no.~7, pp. 1656--1671, 2018.

\bibitem{ren2015faster}
S.~Ren, K.~He, R.~Girshick, and J.~Sun, ``Faster r-cnn: Towards real-time
  object detection with region proposal networks,'' in \emph{Advances in neural
  information processing systems}, 2015, pp. 91--99.

\bibitem{DBLP:conf/iccv/Girshick15}
R.~B. Girshick, ``Fast {R-CNN},'' in \emph{{ICCV}}, 2015, pp. 1440--1448.

\bibitem{DBLP:conf/iccv/HeGDG17}
K.~He, G.~Gkioxari, P.~Doll{\'{a}}r, and R.~B. Girshick, ``Mask {R-CNN},'' in
  \emph{{ICCV}}, 2017, pp. 2980--2988.

\bibitem{DBLP:conf/eccv/LiuAESRFB16}
W.~Liu, D.~Anguelov, D.~Erhan, C.~Szegedy, S.~E. Reed, C.~Fu, and A.~C. Berg,
  ``{SSD:} single shot multibox detector,'' in \emph{{ECCV} {(1)}}, ser.
  Lecture Notes in Computer Science, vol. 9905.\hskip 1em plus 0.5em minus
  0.4em\relax Springer, 2016, pp. 21--37.

\bibitem{DBLP:journals/tmm/LiLLWXFY18}
J.~Li, X.~Liang, J.~Li, Y.~Wei, T.~Xu, J.~Feng, and S.~Yan, ``Multistage object
  detection with group recursive learning,'' \emph{{IEEE} Trans. Multimedia},
  vol.~20, no.~7, pp. 1645--1655, 2018.

\bibitem{DBLP:conf/eccv/BrostowSFC08}
G.~J. Brostow, J.~Shotton, J.~Fauqueur, and R.~Cipolla, ``Segmentation and
  recognition using structure from motion point clouds,'' in \emph{{ECCV}
  {(1)}}, ser. Lecture Notes in Computer Science, vol. 5302.\hskip 1em plus
  0.5em minus 0.4em\relax Springer, 2008, pp. 44--57.

\bibitem{DBLP:journals/ijcv/LadickySRSBCT12}
L.~Ladicky, P.~Sturgess, C.~Russell, S.~Sengupta, Y.~Bastanlar, W.~F. Clocksin,
  and P.~H.~S. Torr, ``Joint optimization for object class segmentation and
  dense stereo reconstruction,'' \emph{Int. J. Comput. Vis.}, vol. 100, no.~2,
  pp. 122--133, 2012.

\bibitem{DBLP:conf/eccv/MunozBH12}
D.~Munoz, J.~A. Bagnell, and M.~Hebert, ``Co-inference for multi-modal scene
  analysis,'' in \emph{{ECCV} {(6)}}, ser. Lecture Notes in Computer Science,
  vol. 7577.\hskip 1em plus 0.5em minus 0.4em\relax Springer, 2012, pp.
  668--681.

\bibitem{DBLP:conf/cvpr/AnguelovTCKGHN05}
D.~Anguelov, B.~Taskar, V.~Chatalbashev, D.~Koller, D.~Gupta, G.~Heitz, and
  A.~Y. Ng, ``Discriminative learning of markov random fields for segmentation
  of 3d scan data,'' in \emph{{CVPR} {(2)}}, 2005, pp. 169--176.

\bibitem{DBLP:conf/cvpr/PohlenHML17}
T.~Pohlen, A.~Hermans, M.~Mathias, and B.~Leibe, ``Full-resolution residual
  networks for semantic segmentation in street scenes,'' in \emph{{CVPR}},
  2017, pp. 3309--3318.

\bibitem{deng2009imagenet}
J.~Deng, W.~Dong, R.~Socher, L.-J. Li, K.~Li, and L.~Fei-Fei, ``Imagenet: A
  large-scale hierarchical image database,'' in \emph{Computer Vision and
  Pattern Recognition, 2009. CVPR 2009. IEEE Conference on}.\hskip 1em plus
  0.5em minus 0.4em\relax IEEE, 2009, pp. 248--255.

\bibitem{yang2011regionwise}
M.~Y. Yang and W.~F{\"o}rstner, ``Regionwise classification of building facade
  images,'' in \emph{Photogrammetric image analysis}.\hskip 1em plus 0.5em
  minus 0.4em\relax Springer, 2011, pp. 209--220.

\bibitem{ecpdataset}
\BIBentryALTinterwordspacing
O.~Teboul, ``Ecole centrale paris facades database.'' [Online]. Available:
  \url{http://vision.mas.ecp.fr/Personnel/teboul/data.php}
\BIBentrySTDinterwordspacing

\bibitem{gadde2016learning}
R.~Gadde, R.~Marlet, and N.~Paragios, ``Learning grammars for
  architecture-specific facade parsing,'' \emph{International Journal of
  Computer Vision}, vol. 117, no.~3, pp. 290--316, 2016.

\bibitem{DBLP:conf/cvpr/CohenSP14}
A.~Cohen, A.~G. Schwing, and M.~Pollefeys, ``Efficient structured parsing of
  facades using dynamic programming,'' in \emph{{CVPR}}, 2014, pp. 3206--3213.

\bibitem{DBLP:conf/3dim/CohenOLP17}
A.~Cohen, M.~R. Oswald, Y.~Liu, and M.~Pollefeys, ``Symmetry-aware fa{\c{c}}ade
  parsing with occlusions,'' in \emph{3DV}, 2017, pp. 393--401.

\end{thebibliography}
\end{document}